\documentclass[a4paper,fleqn]{cas-dc}

\usepackage[authoryear,longnamesfirst]{natbib}
\usepackage{times}
\usepackage{epsfig}
\usepackage{graphicx}
\usepackage{amsmath}
\usepackage{amssymb}
\usepackage{bm}
\usepackage{booktabs}
\usepackage[switch]{lineno}
\usepackage[switch]{lineno}
\usepackage{tabularx}
\usepackage{booktabs}
\usepackage{microtype}
\usepackage{graphicx}
\usepackage{algorithm}
\usepackage{algorithmic}
\usepackage{subcaption}
\usepackage{multirow} 
\usepackage{array,longtable,booktabs,ragged2e}
\usepackage{caption}
\usepackage{changepage}
\usepackage{xcolor}
\usepackage{listings}
\usepackage{enumitem}
\usepackage{listings}
\usepackage{xcolor}  
\usepackage{makecell}
\usepackage{tabularx}
\usepackage{threeparttable}
\usepackage{placeins}

\newcommand{\rev}[1]{\texorpdfstring{\textcolor{blue}{#1}}{#1}}

\lstdefinestyle{json}{
    language=JavaScript,                 
    morekeywords={true,false,null},      
    keywordstyle=\color{blue!90!black}\bfseries,
    stringstyle=\color{orange!90!black},
    commentstyle=\color{gray},
    basicstyle=\ttfamily\small,
    columns=flexible,
    keepspaces=true,
    showstringspaces=false,
    backgroundcolor=\color{gray!5},    
    frame=tb,                            
    rulecolor=\color{gray!50},
}

\usepackage{balance}
\graphicspath{{./image/}}
\def\tsc#1{\csdef{#1}{\textsc{\lowercase{#1}}\xspace}}
\tsc{WGM}
\tsc{QE}

\begin{document}
\let\WriteBookmarks\relax
\def\floatpagepagefraction{1}
\def\textpagefraction{.001}

    \shorttitle{Improving Building Energy Efficiency with an Indoor Environmental World Model and Counterfactual Adaptive Learning}

\shortauthors{Xu Yang et al.}

\title[mode=title]{Improving Building Energy Efficiency with an Indoor Environmental World Model and Counterfactual Adaptive Learning}



%



\author[1]{Xu Yang}[type=editor,style=chinese, orcid=0009-0008-6281-1606]
\ead{yangxu24@mails.tsinghua.edu.cn}


\author[2,3]{Kailai Sun}[type=editor,style=chinese,orcid=0000-0003-1648-3409]
\cormark[1]

\ead{skl24@mit.edu}

\author[1]{Dianyu Zhong}[type=editor,style=chinese,orcid=0000-0002-3262-4905]

\author[1]{Irfan Qaisar}[type=editor,style=chinese, orcid=0000-0002-4831-977X]

\author[4]{Yu Yang}[type=editor,style=chinese, orcid=]
\ead{yangyu21@xjtu.edu.cn}

\author[1]{Qianchuan Zhao}[type=editor,style=chinese,orcid=0000-0002-7952-5621]
\cormark[1]


\ead{zhaoqc@mail.tsinghua.edu.cn}

\author[5]{Jian Kang}[type=editor,style=chinese,orcid=0000-0001-8995-5636]
\ead{j.kang@ucl.ac.uk}



\affiliation[1]{organization={Center for Intelligent and Networked Systems, Department of Automation, BNRist, Tsinghua University},
            city={Beijing},
            postcode={100084}, 
            country={China}}


\affiliation[2]{organization={Singapore-MIT Alliance for Research and Technology Centre (SMART), Massachusetts Institute of Technology},
            postcode={138602}, 
            country={Singapore}}
\cortext[1]{Corresponding authors.}

\affiliation[3]{organization={Urban Mobility Lab, Massachusetts Institute of Technology},
            addressline={Cambridge}, 
            state={MA 02139},
            country={United States}}

\affiliation[4]{organization={School of Automation Science and Engineering, Xi'an Jiaotong University},
            city={Xi'an, Shaanxi},
            postcode={710049}, 
            country={China}}

\affiliation[5]{organization={Institute for Environmental Design and Engineering, The Bartlett, University College London},
            city={London},
            postcode={WC1E 6BT}, 
            country={United Kingdom}}






\maketitle

\begin{abstract}
    Optimizing indoor climate systems plays a critical role in achieving the United Nations' Sustainable Development Goals 11 and 13. However, indoor delayed thermodynamic responses and partial observability severely hinder existing reinforcement learning (RL) methods, which are primarily limited by implicit thermal inertia, occupancy dynamic prediction, and cumulative prediction errors during long-horizon rollouts. Thus, we propose a novel occupant-centric indoor climate control framework. It develops indoor environmental world models to predict future physical thermal baselines and dynamic occupancy, explicitly capturing latent thermal inertia. By formulating adaptive regulation as counterfactual advantage learning against future environments via a Branching Dueling Q-Network, and integrating a Held-Action TD-$\lambda$ target, it effectively mitigates delayed credit assignment. Additionally, a proposed physics-aligned smoothness regularization suppresses high-frequency actuator fluctuations, potentially extending equipment service life. Experiments on SemiBuildingSim and Sinergym demonstrate that our method achieves a superior multi-objective trade-off.  Compared to the best-performing methods, it further reduces energy consumption by 11.2\%, occupant discomfort by 44.2\%, and action fluctuation by 47.3\%, achieving better sample efficiency and cross-region effectiveness. This interdisciplinary study offers an energy-efficient, occupant-centric AI solution for control science, building science, and energy sustainability.
\end{abstract}

\section{Introduction}
\label{section1}

Energy-efficient buildings are essential for improving energy utilization and supporting sustainable economic development. 
According to the Intergovernmental Panel on Climate Change and International Energy Agency, achieving net-zero carbon emissions by 2050 is critical for humanity \cite{lee2023climate, unfccc2023gst}, especially as global average temperatures exceeded the 1.5°C warming threshold in 2024 for the first time~\cite{wmo2025,tollefson2025earth}.
Artificial intelligence (AI) is emerging as a promising approach to enhancing building energy efficiency and supporting the low-carbon transition\cite{luo2019development, kaack2022aligning}.
Buildings account for approximately 32\% of global energy consumption and 34\% of energy-related CO$_2$ emissions \cite{2022un}, with heating, ventilation, and air conditioning (HVAC) systems contributing the largest proportion of operational energy consumption \cite{perez2008hvac}.
Recent studies further highlight the substantial potential for building and urban energy \cite{sun2026sense} optimization, particularly in light of the increasing cooling demand driven by climate change~\cite{creutzig2022demand}.

On the other hand, occupants spend nearly 90\% of their time indoors \cite{EPA_indoor}, making indoor environmental quality particularly important. 
Thermal comfort, as defined in ASHRAE Standard 55 \cite{ashrae2023standard55, fanger1970thermal}, is closely associated with occupants' health, satisfaction, and productivity. 
Consequently, intelligent HVAC control requires a coordinated optimization of energy consumption, occupant comfort and physical system consistency \cite{wang2021reinforcement, hanafi2024advancing}.
Occupant-centric control is becoming a representative AI-for-sustainability solution, which contributes to the United Nations Sustainable Development Goals (SDGs), particularly SDG 11 (Sustainable Cities and Communities) and SDG 13 (Climate Action) \cite{wang2023occupant,  fuso2019connecting}.

A practical challenge in HVAC control is the delayed thermodynamic response. 
Latent heat stored in building structures and water loops is only partially captured by sensors, while this thermal inertia causes the effects of control actions to emerge gradually over multiple time intervals \cite{verbeke2018thermal,glashier2024iterative}.
The resulting partial observability and delayed credit assignment require controllers to account for accumulated latent heat alongside current observations.
Furthermore, dynamic occupancy patterns continuously affect internal heat gains and ventilation demands, introducing substantial nonstationarity into the indoor environment \cite{nagy2023ten}.

Model Predictive Control (MPC) has been regarded as a classic approach for HVAC control~\cite{ bayasgalan2024comprehensive, dong2026optimal}. 
However, MPC typically requires formulating thermodynamic equations and calibrating model parameters for each building, making accurate prediction models difficult to obtain~\cite{ren2026agenticontrol}.
\textcolor{black}{Additionally, online optimization guided by the prediction model could be sensitive to model mismatch, forecast errors, constraint tuning, and solver overhead in real deployments \cite{wang2021reinforcement}.} 
Model-free Deep Reinforcement Learning (DRL) offers a more flexible alternative by learning control policies directly from interaction data without requiring calibrated dynamics models \cite{haarnoja2018soft}.
Yet in systems with delayed thermal responses, standard temporal-difference learning can struggle to assign credit to actions whose effects become apparent later.
Stacking past observations or using recurrent policies partly address partial observability~\cite{hausknecht2015deep,ni2023transformers}, but the policy or value function still infer latent thermal drift implicitly from observation histories. Moreover, these standard model-free approaches do not explicitly predict occupancy dynamics, limiting their ability to anticipate and respond to occupancy-driven thermal variations.

Recent advances in world models offer a promising approach to learning environment dynamics from data \cite{lecun2022path, ding2025understanding}.
Methods such as Dreamer \cite{hafner2023mastering}, TD-MPC \cite{hansen2022temporal}, and Model-Based Value Expansion \cite{feinberg2018model, buckman2018sample} use learned dynamics to support planning or value estimation.
However, directly rolling out candidate future action sequences with world models may not always be suitable for HVAC control.
Such rollouts could be computationally expensive in systems with multiple actuators, while model errors accumulate over long prediction horizons \cite{janner2019when, kurutach2018model}.
More importantly, in thermodynamic systems with delayed dynamics,
the primitive challenge lies in predicting how the system evolves under current control 
rather than focusing solely on roll-outing future action sequences.
This suggests a different role for world models: estimating a short-horizon thermal baseline that reflects latent building inertia.

Motivated by this insight, we propose a framework that integrates indoor environmental world models with counterfactual adaptive learning for empowering building energy efficiency.
To address partial observability and thermal inertia, the framework uses a MambaFormer-based~\cite{xu2024integrating} indoor environmental world model to predict a delayed thermal baseline.
We integrate this baseline with a Branching Dueling Q-Network (BDQ) \cite{tavakoli2018action} for counterfactual adaptive learning,
evaluating candidate actions compared to maintaining the current action.
To address delayed credit assignment, we formulate a Held-Action TD-$\lambda$ target that efficiently propagates the short-horizon consequences of the executed action.
We further introduce a physics-aligned smoothness reward to reduce high-frequency actuator fluctuations and promote physically plausible operation.
The proposed framework is trained and evaluated on the custom-built calibrated real-world simulator, SemiBuildingSim~\cite{xing2022honeycomb,zhong2025topology}, and the Sinergym~\cite{campoy2025sinergym} benchmark.

The main contributions of this study are as follows.
\begin{enumerate}[leftmargin=*]
    \item 
    We develop an indoor environmental world model (IEWM) that
    leverages interaction data to predict the delayed thermal response, providing a thermal baseline for downstream control tasks.
    
    \item 
    We propose counterfactual adaptive learning (CAL), which uses
    the predicted thermal baseline to estimate the counterfactual advantage
    of each candidate action.
    CAL incorporates a Held-Action TD-$\lambda$ target for delayed credit
    assignment and a physics-aligned reward for smooth actuation.
    
    \item 
    Experiments demonstrate that our framework achieves superior multi-objective control performance.
    Relative to strong baselines, our framework reduces HVAC energy consumption by \textbf{11.2\%}, occupant discomfort by \textbf{44.2\%}, and action fluctuation by \textbf{47.3\%},
    while maintaining performance across different regions.
\end{enumerate}

Taken together, this study addresses the following overall research question: given indoor delayed thermal response, can a domain-specific world model learn the future physical evolution of the thermal environment from partial observation trajectories, and how its prediction serve as a reliable thermal baseline for energy-efficient indoor control?
Section~\ref{sec:approach} explains how the study is designed to answer this question, and Section~\ref{sec:results} answers it through four sub-questions.

The remainder of this paper is organized as follows.
Section~\ref{sectionL} reviews related work.
Section~\ref{sec:approach} presents the overall methodological approach.
Section~\ref{sec:method} describes the model development.
Section~\ref{sec:exp_env_details} introduces the experimental design and evaluation environments.
Section~\ref{sec:results} reports the results and the ablation study.
Section~\ref{sec:discussion} discusses the improvement and broader impact of this research, and Section~\ref{sec:conclusion} concludes the paper.

\section{Related Work}
\label{sectionL}
\subsection{Reinforcement learning applications in HVAC systems}
RL has become a promising alternative for HVAC control as it learns policies directly from interaction data without requiring a calibrated dynamics model \cite{van2016deep, haarnoja2018soft,  wang2025safe}. 
Prior work has applied RL to HVAC control, residential demand response, and building energy optimization \cite{vazquez2019reinforcement, wang2021reinforcement, nguyen2024modelling, sun2024individual, wang2024experimental, togashi2025reward}. 
Existing EnergyPlus-based simulators and benchmarks (e.g., Sinergym) have improved reproducible evaluation \cite{crawley2001energyplus, jimenez2021sinergym}. But energy-efficient HVAC control remains challenging as hidden thermal storage creates delayed action effects under partial observability. 
Existing solutions often rely on observation histories or modern memory-augmented architectures such as recurrent networks and transformers \cite{lu2024rethinking, zengpoliformer}, but they do not explicitly expose the policy to the future drift induced by latent thermal inertia. 
In contrast, IEWM models this drift as a thermal baseline, feeds it into a branching action-value architecture \cite{tavakoli2018action}, and learns a Counterfactual Advantage with a Held-Action TD-$\lambda$ target.

\subsection{Occupant-centric HVAC control}
Occupant-centric HVAC control integrates real-time indoor sensing, such as environmental quality, occupant information, and occupant-building interactions, to dynamically optimize energy efficiency and occupant comfort \cite{fanger1970thermal,ashrae2023standard55}. 
It has motivated many building-control studies that explicitly integrate comfort metrics, occupancy information \cite{zhang2026physics}, and weather information into decision making \cite{ yuan2024review, liang2025human, meimand2024personal}. MPC has been widely adopted, but its implementation typically depends on accurate system identification, careful calibration, and repeated online optimization \cite{drg2020review}. 
Our method addresses a similar occupant-centric objective by learning an indoor environmental world model, which jointly forecasts the thermal baseline and occupancy dynamics.

\subsection{World models for reinforcement learning}
The integration of learned dynamics and reinforcement learning dates back to Dyna-style architectures \cite{sutton1991dyna} and has evolved into modern world-model methods that imagine trajectories in latent space \cite{ding2025understanding}. In continuous control, learned dynamics have been used for sample-efficient policy optimization and value expansion through PETS, MBPO, and their extensions \cite{chua2018deep, buckman2018sample, janner2019when}. Recently, Dreamer, MuZero, and TD-MPC have further established scalable world models for long-horizon planning \cite{hansen2022temporal, hafner2025mastering, hansen2024tdmpc2}. Furthermore, the paradigm has rapidly advanced toward foundation-level predictive architectures (e.g., JEPA), generative interactive environments, and training methods utilizing verifiable rewards \cite{bardes2024vjpea, bruce2024genie, wu2025rlvr, zhao2025generalist, georgiev2025pwm}. 
At a broader level, predictive world models have been argued to be central to generally capable intelligent systems \cite{lecun2022path}, while long-sequence architectures such as Mamba improve temporal modeling over extended horizons \cite{gu2023mamba}. 
Our method differs from these methods in how the model is used: instead of exhaustive action-sequence planning, the world model is used to estimate a thermal momentum trajectory under a held-action protocol, forming a counterfactual baseline for action evaluation; during training, it further provides short held-action rollouts for TD-$\lambda$.

\begin{figure*}[ht]
    \centering
    \includegraphics[width=1.0\linewidth]{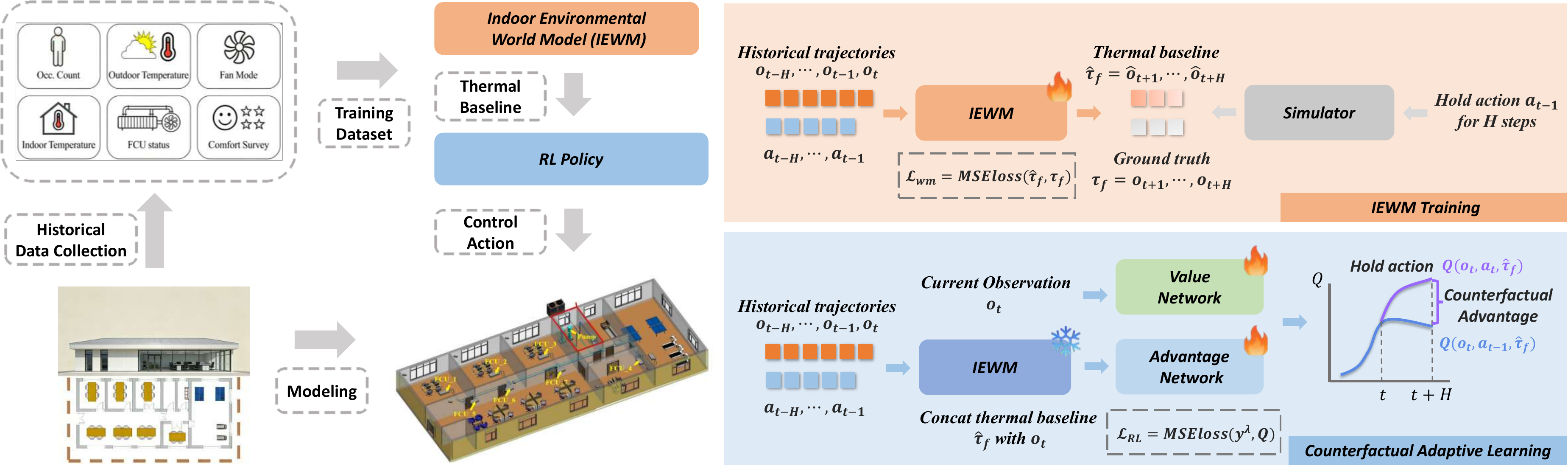}
    \caption{Overall methodology and proposed framework establishment. Left: research workflow, linking historical data collection and building simulation to prediction and control. Right: the framework is trained in two phases. \textbf{Phase I:} IEWM training. \textbf{Phase II:} counterfactual adaptive learning (CAL). The IEWM is frozen during Phase II.}
    \label{fig:framework}
\end{figure*}

\section{Overall Methodology}
\label{sec:approach}

The study starts from the view that a building is a delayed thermal system.
Current measurements only partially reveal the heat stored in building structures and water loops. 
This stored heat causes the effects of past control actions on indoor temperature to unfold over several time steps, while heat gains from occupants further affect the indoor temperature.
This view determines the factors built into the model.
We therefore train a world model to jointly predict occupancy dynamics and indoor thermal evolution from histories of observations and control actions. 
By capturing the indoor thermal dynamics, the model provides an explicit thermal baseline for the controller’s next decision.
For the input of IEWM, past actuator commands help the model infer delayed heating and cooling effects, while occupancy and outdoor weather account for changes in thermal loads. 
The slow response also motivates smoothness regularization, since frequent actuator switching can increase energy use and equipment wear while having little immediate effect on indoor temperature.

Motivate by these insights, we design the control framework around counterfactual advantage, evaluating each candidate action compared to maintaining the current actuator actions. 
To establish this reference, we first train IEWM to predict a thermal baseline (Phase I), which represents how the indoor environment would evolve if the most recently applied actions were held fixed, while stored heat, occupancy, and outdoor conditions continue to affect its evolution.
CAL uses the predicted thermal baseline as a shared input to its branch-wise advantage streams, which estimate counterfactual advantages for candidate actions against this thermal baseline (Phase II).
During training, the Held-Action TD-$\lambda$ target uses short predicted trajectories following the executed action to account for thermal effects that emerge over several control intervals.
The predicted thermal baseline thus serves as a reference for action selection and helps the controller learn the delayed effects of its actions (right panel of Fig.~\ref{fig:framework} and Section~\ref{sec:method}).

The left panel of Fig.~\ref{fig:framework} shows how we build and evaluate the framework. 
We use office records of occupant counts, indoor and outdoor temperatures, and fan operating modes, together with the building and HVAC layouts, to construct the office simulation.
We train the IEWM on trajectories generated by the simulator and evaluate the policy in closed loop by applying each selected action and feeding the resulting observations back to the policy. 
We assess thermal comfort according to ASHRAE Standard 55~\cite{ashrae2023standard55}.

In the office, we measure prediction accuracy and compare the policy with MPC and reinforcement learning baselines in terms of energy use, occupant comfort, and actuator fluctuation. 
We examine how the smoothness penalty affects the balance among these objectives, whether the method remain effectiveness under long-horizon occupant-centric control, and how the prediction contributes to policy training and action scoring. 
To test whether the framework applies under different building types and regions , we also evaluate it on a data center with a different HVAC system and climate, measuring prediction accuracy, energy use, and temperature violations.(Sections~\ref{sec:exp_env_details} and~\ref{sec:results}).

\section{Framework Establishment}
\label{sec:method}
This section describes the specific method and model development shown on the right of Fig.~\ref{fig:framework}.
Our framework integrates an indoor environmental world model (IEWM) with counterfactual adaptive learning (CAL).
The IEWM captures the latent building thermal inertia and predicts a delayed thermal baseline. CAL combines this baseline with a modified BDQ Network~\cite{tavakoli2018action} for counterfactual action evaluation.
To address delayed credit assignment, we introduce a Held-Action TD-$\lambda$ target.
We also incorporate a physics-aligned smoothness reward to reduce actuator fluctuations.

\subsection{Thermodynamic Problem Formulation}
\label{subsec:pomdp}
We model the optimal HVAC control problem as a Partially Observable Markov Decision Process (POMDP), defined by the tuple $\mathcal{M} = (\mathcal{S}, \mathcal{O}, \mathcal{A}, P, R, \gamma)$.

In building energy dynamics, the true underlying physical state $\mathcal{S}$ includes unmeasurable thermal storage in structural mass (e.g., concrete walls and furniture). However, the agent only receives partial sensory observations at time step $t$, $o_t \in \mathcal{O}$:
\begin{equation}
    o_t = [T^{air}_t, H_t, O_t, T^{out}_t, S_t, Z_t],
\end{equation}        
where $T^{air}_t$ is indoor air temperature, $H_t$ is relative humidity, $O_t$ denotes measured or estimated occupancy pattern, $T^{out}_t$ denotes outdoor temperature, $S_t$ denotes solar radiation intensity, and $Z_t$ collects auxiliary HVAC measurements.

The controller maps observations to a multidimensional discrete action vector $a_t \in \mathcal{A}$, covering discretized HVAC actuator configurations such as fan mode or set temperature. The physical transition $P(s_{t+1}|s_t,a_t)$ is governed by thermodynamic dynamics. Because structural thermal mass stores and releases heat slowly, the observed process $P(o_{t+1}|o_t,a_t)$ exhibits delayed and partially observed behavior: the effect of an action at time $t$ may only become clearly visible after several control intervals. The objective is to optimize a policy $\pi$ that maximizes the discounted return:
$
    \max_{\pi} \mathbb{E}_{\pi}\left[\sum_{t=0}^{\infty} \gamma^t r_t \right].
$

\subsection{Indoor Environmental World Model (IEWM)}
\label{subsec:iewm}

To capture the system's latent thermal inertia, we train an autoregressive world model $f_\theta$ as a surrogate thermodynamic simulator. Unlike conventional model-based RL methods (e.g. DreamerV3~\cite{hafner2025mastering}) that condition on candidate actions for planning, our world model is utilized to predict the future indoor thermal baseline with held action.

Specifically, we formulate the forward projection on the historical observation  $o_{t-N:t}$ and the prior action sequence $a_{t-N:t-1}$. The resulting baseline rollout over horizon $H$ is defined as
\begin{equation}
    \hat{\tau_f} = \hat{o}_{t+1:t+H}
    =
    f_\theta\!\left(
    o_{t-N:t}, a_{t-N:t-1}
    \;\middle|\;
    a_{t:t+H-1}=a_{t-1}
    \right).
\end{equation}
By assuming held-action, the predicted trajectory $\hat{\tau_f}$ serves as a counterfactual baseline that captures the building's latent thermal inertia when the previous action $a_{t-1}$ is maintained. To train this model consistently, we enforce a held-action protocol ($a_{t:t+H-1} = a_{t-1}$) during offline dataset collection, allowing the realized sequence $o_{t+1:t+H}$ to act as the direct, unbiased supervision target. The parameterized sequence model $f_\theta$ is then optimized via multi-step MSE loss:
\begin{equation}
    \mathcal{L}_{wm}(\theta) = \sum_{k=1}^{H} \| o_{t+k} - \hat{o}_{t+k} \|_2^2.
\end{equation}
Minimizing this loss allows the world model to capture delayed thermal response patterns directly from data, without requiring manual derivation of a white-box building model. 
Most importantly, while the $H$-step future supervision target is rigorously constrained to be constant-action, the historical input window $(o_{t-N:t}, a_{t-N:t-1})$ remains entirely unconstrained. This structural training asymmetry naturally resolves the counterfactual challenge: it trains the sequence model to predict the thermal baseline starting from any dynamically varying history. 

\subsection{Counterfactual Adaptive Learning (CAL)}
\label{subsec:cal}

Given the predicted baseline $\hat{\tau_f}$, we use a modified Branching Dueling Q-Network (BDQ) to handle multidimensional discrete HVAC control. Since real-world actuators are continuous but operationally quantized, we discretize each action dimension into bins $|\mathcal{A}_d|$ and model branch-wise action values for each actuator dimension $d \in \{1,\dots,D\}$:
\begin{equation}
    Q_d(o_t, a_{t,d}, \hat{\tau_f})
=
V_\psi(o_t)
+
A_{\eta_d}(o_t, a_{t,d}, \hat{\tau_f})
-
\max_{a' \in \mathcal{A}_d}
A_{\eta_d}(o_t, a', \hat{\tau_f}).
\end{equation}
The value stream $V_\psi(o_t)$ evaluates the current physical state using immediate observations only, while the advantage stream $A_{\eta_d}(o_t, a_{t,d}, \hat{\tau_f})$ conditions on the predicted thermal baseline. This decoupling is intended to separate instantaneous state evaluation from action evaluation under delayed thermal drift.

To formalize the counterfactual interpretation, we define the reference held-action for each branch $d$ as
$
a^{\mathrm{held}}_{t,d} = a_{t-1,d},
$
and formulate the branch-wise counterfactual advantage relative to this baseline as
\begin{equation}
\begin{aligned}
\Delta_d(o_t, a_{t,d}, \hat{\tau_f})
&= Q_d(o_t, a_{t,d}, \hat{\tau_f}) 
 - Q_d(o_t, a^{\mathrm{held}}_{t,d}, \hat{\tau_f}) \\
&= A_{\eta_d}(o_t, a_{t,d}, \hat{\tau_f}) 
- A_{\eta_d}(o_t, a^{\mathrm{held}}_{t,d}, \hat{\tau_f}).
\end{aligned}
\label{eq:cf_adv}
\end{equation}
Under this definition, the agent explicitly evaluates candidate actions against the baseline of maintaining the previous action. Thus, the learned action preference is directly driven by the relative advantage over the thermal baseline encoded by $\hat{\tau_f}$.

Under the BDQ factorization, greedy action selection decomposes across branches, which makes large multidimensional discrete action spaces tractable:
\begin{equation}
    a_{t,d}^* = \arg\max_{a_{t,d}} Q_d(o_t, a_{t,d}, \hat{\tau_f}) \equiv \arg\max_{a_{t,d}}\Delta_d(o_t, a_{t,d}, \hat{\tau_f}).
\end{equation}
Equivalently, since $Q_d(o_t, a^{\mathrm{held}}_{t,d}, \hat{\tau_f})$ is constant with respect to the candidate actions, the same action can be selected by maximizing $\Delta_d(o_t, a_{t,d}, \hat{\tau_f})$.

\subsection{Delay-Aware Credit Assignment via Held-Action TD-$\lambda$ Target}
\label{subsec:tdlambda}

A key challenge in HVAC control is that the thermal response of an actuator action manifests only after several control intervals. To improve long-horizon credit assignment under these delayed dynamics, we augment TD learning with a lookahead return target.

During online interaction, the selected action $a_t$ is first executed in the real environment to obtain the ground truth one-step transition $(o_{t+1}, r_t)$. Starting from this next observation, we query the world model under the held action protocol to approximate the short-term thermal response:
\begin{equation}
    \hat{\tau_f}' = \hat{o}_{t+2:t+H+1}^{(a_t)}
    =
    f_\theta\!\left(
    o_{t-N+1:t+1}, a_{t-N+1:t}
    \;\middle|\;
    a_{t+1:t+H}=a_t
    \right).
\end{equation}
This gives a hybrid trajectory that concatenates the empirical observation with the model's predictions:
\begin{equation}
    \tilde{o}_{t+1}^{(a_t)} = o_{t+1}, \qquad
    \tilde{o}_{t+2:t+H+1}^{(a_t)} = \hat{o}_{t+2:t+H+1}^{(a_t)}.
\end{equation}
Future intermediate rewards are estimated using a deterministic rule-based reward evaluator $g(\cdot)$:
\begin{equation}
    \hat{r}_{t+k}^{(a_t)} = g\!\left(\tilde{o}_{t+k}^{(a_t)}, a_t\right), \qquad k = 1, \dots, H-1.
\end{equation}

The held-action rollout target serves as a proxy for the true state dynamics, assuming the controller maintains the selected action over the subsequent $H$-step control horizon.
The short-horizon held-action rollout provides a lower-variance estimate of the return while capturing delayed thermal effects beyond the one-step TD horizon. We use the resulting trajectory to construct the \(h\)-step expanded return:

\begin{equation}
    G_t^{(h)}(a_t) = r_t + \sum_{k=1}^{h-1} \gamma^k \hat{r}_{t+k}^{(a_t)} + \gamma^h V_{\bar{\psi}}\left(\tilde{o}_{t+h}^{(a_t)}\right).
\end{equation}
This target combines the empirical one-step reward, the intermediate rewards predicted by the held-action rollout, and a bootstrapped value estimate evaluated at the simulated future state using the target network $V_{\bar{\psi}}$.

To balance short and long-range delayed effects, we combine these multi-step targets into a TD-$\lambda$ mixture:
\begin{equation}
    y_t^\lambda = (1-\lambda) \sum_{h=1}^{H-1} \lambda^{h-1} G_t^{(h)}(a_t) + \lambda^{H-1} G_t^{(H)}(a_t).
    \label{eq: td lambda}
\end{equation}
The BDQ network is then optimized by minimizing:
\begin{equation}
    \mathcal{L}_{RL}(\Phi) = \mathbb{E}_{\mathcal{D}} \left[ \frac{1}{D} \sum_{d=1}^{D} \left( y_t^\lambda - Q_d(o_t, a_{t,d}, \hat{\tau_f}; \Phi) \right)^2 \right].
    \label{eq:loss_rl}
\end{equation}
Overall, the world model serves two complementary functions: it provides the thermal baseline $\hat{\tau_f}$ for evaluating candidate actions, and generates short held-action prediction for delayed credit assignment.

\subsection{Physics-Aligned Smoothness Regularization}
\label{subsec:smooth}

An undesirable behavior frequently observed when applying RL to physical control systems is high-frequency action switching. In HVAC settings, such oscillatory behavior may yield minimal short-term gains while high-frequency adjustments accelerate wear on valves, dampers, and variable-speed equipment, ultimately shortening their lifespan. To encourage physically plausible control, we augment the reward with an action smoothness  regularization.

We consider two formulations for measuring divergence between consecutive actions: a logarithmic reward and a linear $L_1$ reward,
\begin{equation}
\begin{aligned}
R_{smooth_1, t}
&= \sum_{d=1}^{D} \log\big(1 + |a_{t,d} - a_{t-1,d}|\big), \\
R_{smooth_2, t}
&= \sum_{d=1}^{D} |a_{t,d} - a_{t-1,d}|.
\end{aligned}
\end{equation}
The logarithmic form is sub-linear and therefore more tolerant of occasional large adjustments. This prevents the smoothness penalty from dominating the overall reward signal during initial random exploration, thereby stabilizing early-stage training. Whereas the $L_1$ form penalizes action variation proportionally at all scales. The final reward combines comfort, energy, and smoothness:
\begin{equation}
    r_t = - r_{comfort} - \alpha \cdot r_{energy} - \beta \cdot R_{smooth, t},
\end{equation}
where $r_{comfort}$ encodes PPD-based comfort, $r_{energy}$ is instantaneous electrical power consumption, and $\{\alpha, \beta\}$ balance the multi-objective trade-off. Full pseudocode is below.


\begin{algorithm*}[ht]
\caption{Our proposed method: integrating indoor environmental world model with counterfactual adaptive learning}
\label{alg:training}
\begin{algorithmic}[1]

\REQUIRE Indoor environmental world model $f_{\theta}$, rule-based reward evaluator $g$, behavior Q-network $\Phi=\{V_\psi, A_\eta\}$, target network $\bar{\Phi}$, replay buffer $\mathcal{D}$

\STATE \textbf{Phase I: IEWM Training}
\STATE Pretrain IEWM $f_\theta$ on held-action target horizons by minimizing predictive MSE loss.
\STATE \textbf{Phase II: Counterfactual Adaptive Learning}
\STATE Freeze $f_\theta$.
\FOR{each training episode}
    \STATE Reset environment and observe initial state $o_0$
    \STATE Execute random policy for $N$ steps to collect initial history windows $o_{0:N}$ and $a_{0:N-1}$
    \FOR{each timestep $t=N, N+1, \dots, T$}
        \STATE \textit{// 1. Estimate thermal baseline}
        \STATE Predict thermal baseline:
        \STATE \quad $\hat{\tau_f} = f_{\theta}\!\left(o_{t-N:t}, a_{t-N:t-1} \;\middle|\; a_{t:t+H-1}=a_{t-1}\right)$

        \STATE \textit{// 2. Branch-wise action selection via BDQ}
        \STATE Sample $u \sim \mathrm{Uniform}(0,1)$
        \IF{$u < \epsilon$}
            \STATE Sample random exploration action $a_t \in \mathcal{A}$
        \ELSE
            \FOR{each branch $d=1,\dots,D$}
                \STATE Evaluate
                \STATE \quad $Q_d(o_t, a_{t,d}, \hat{\tau_f}) = V_\psi(o_t) + A_{\eta_d}(o_t, a_{t,d}, \hat{\tau_f}) - \max_{a' \in \mathcal{A}_d} A_{\eta_d}(o_t, a', \hat{\tau_f})$
                \STATE Select greedy action $a_{t,d} = \arg\max_{a \in \mathcal{A}_d} Q_d(o_t, a, \hat{\tau_f}; \Phi)$
            \ENDFOR
        \ENDIF

        \STATE \textit{// 3. One-step execution and committed rollout construction}
        \STATE Execute $a_t$ for one environment step, observe $o_{t+1}$ and $r_t$
        \STATE Query frozen world model under held-action condition:
        \STATE \quad $\hat{\tau_f}'=\hat{o}_{t+2:t+H+1}^{(a_j)} = f_\theta\!\left(o_{t-N+1:t+1}, a_{t-N+1:t} \;\middle|\; a_{t+1:t+H}=a_t\right)$
        \STATE Store augmented transition $(o_{t}, a_{t}, r_t, o_{t+1}, \hat{\tau_f}, \hat{\tau_f}')$ in $\mathcal{D}$

        \STATE \textit{// 4. Held-Action TD-$\lambda$ update}
        \IF{time to update parameters}
            \STATE Sample a mini-batch $\mathcal{B}$ from $\mathcal{D}$
            \FOR{each sampled transition indexed by $j$ in $\mathcal{B}$}
                \STATE Extract held-action thermal baseline $\hat{\tau_f}'=\hat{o}_{t+2:t+H+1}^{(a_j)}$ 
                \STATE Compute imagined rewards $\hat{r}_{j+k}=g(\tilde{o}_{j+k}^{(a_j)}, a_j)$ for $k \in \{1,\dots,H-1\}$
                \STATE Compute Held-Action TD-$\lambda$ target $y_j^\lambda$ as Eq.~\ref{eq: td lambda}
            \ENDFOR
            \STATE Compute batch loss $\mathcal{L}_{RL}(\Phi)=\frac{1}{D}\sum_{d=1}^{D}(y_j^\lambda-Q_d)^2$
            \STATE Update behavior network $\Phi$ via gradient descent
            \STATE Soft update target parameters: $\bar{\Phi} \leftarrow \kappa \Phi + (1-\kappa)\bar{\Phi}$
        \ENDIF
    \ENDFOR
\ENDFOR

\end{algorithmic}
\end{algorithm*}

\section{Experimental Design}
\label{sec:exp_env_details}

The framework is evaluated in two buildings located in different climate regions to examine its cross-region (China and United States) effectiveness for building energy efficiency.
SemiBuildingSim (Section~\ref{subsec:semibuildingsim}) is a seven-zone office building in Hebei, China, calibrated with real operational data and including occupant number and activity in each zone. It is used to verify the prediction accuracy of the IEWM and the occupant-centric control performance of CAL, including the effects of smoothness regularization and long-horizon prediction.
Sinergym \texttt{2ZoneDataCenterHVAC} (Section~\ref{subsec:sinergym_env}) is a data center under Chicago weather in the United States, with a different building use, HVAC system and control interval. It is used to verify that the prediction and energy-saving performance of the framework remain effective in different regions and building types.

\subsection{SemiBuildingSim Environment}
\label{subsec:semibuildingsim}

\subsubsection{Environment Construction and Case Study Modeling}
\label{subsubsec:case_study_setup}

To comprehensively evaluate the proposed method, we utilize SemiBuildingSim, a high-fidelity simulator designed to replicate the complex thermal and energy dynamics of a real-world building~\cite{zhong2025topology}. The simulation environment is grounded in historical data collected from a seven-zone office building in Hebei Province, China~\cite{xing2022honeycomb}. The building operates primarily during standard working hours (9:00- 19:00 on weekdays), with reduced occupancy around lunchtime. An average of 12.4 people occupied the office (August 7 to August 22), with typical weekday usage and no regular weekend activity.

\begin{figure*}[htbp]
    \centerline{
    \includegraphics[width=\columnwidth, trim=0cm 0cm 0cm 0cm, clip]{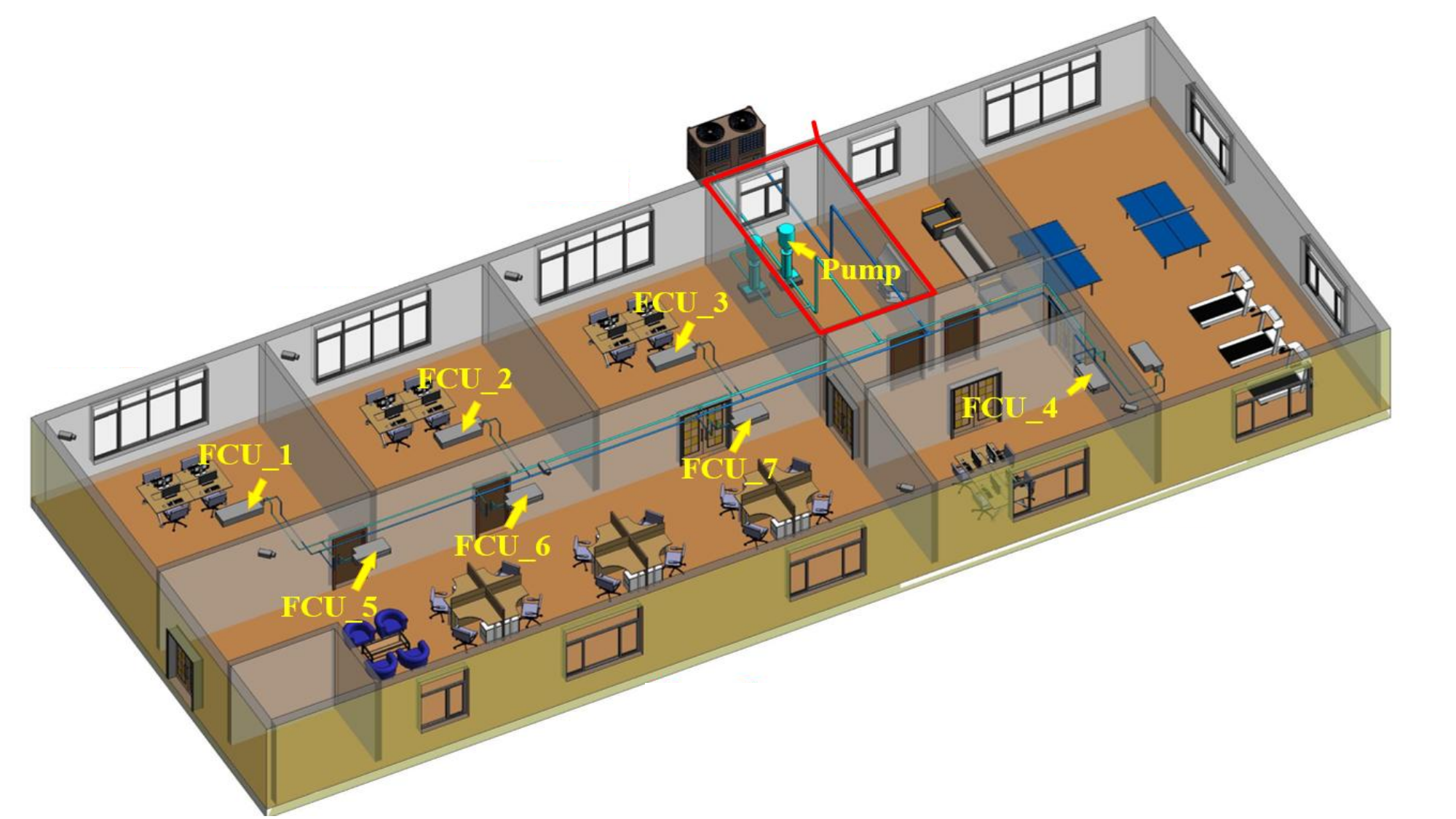}
    }    
    \caption{Layout of the office building \cite{zhong2025topology}.}   
    \label{fig:layout}
\end{figure*}

\begin{figure*}[htbp]
    \centering
    \includegraphics[width=1.0\linewidth]{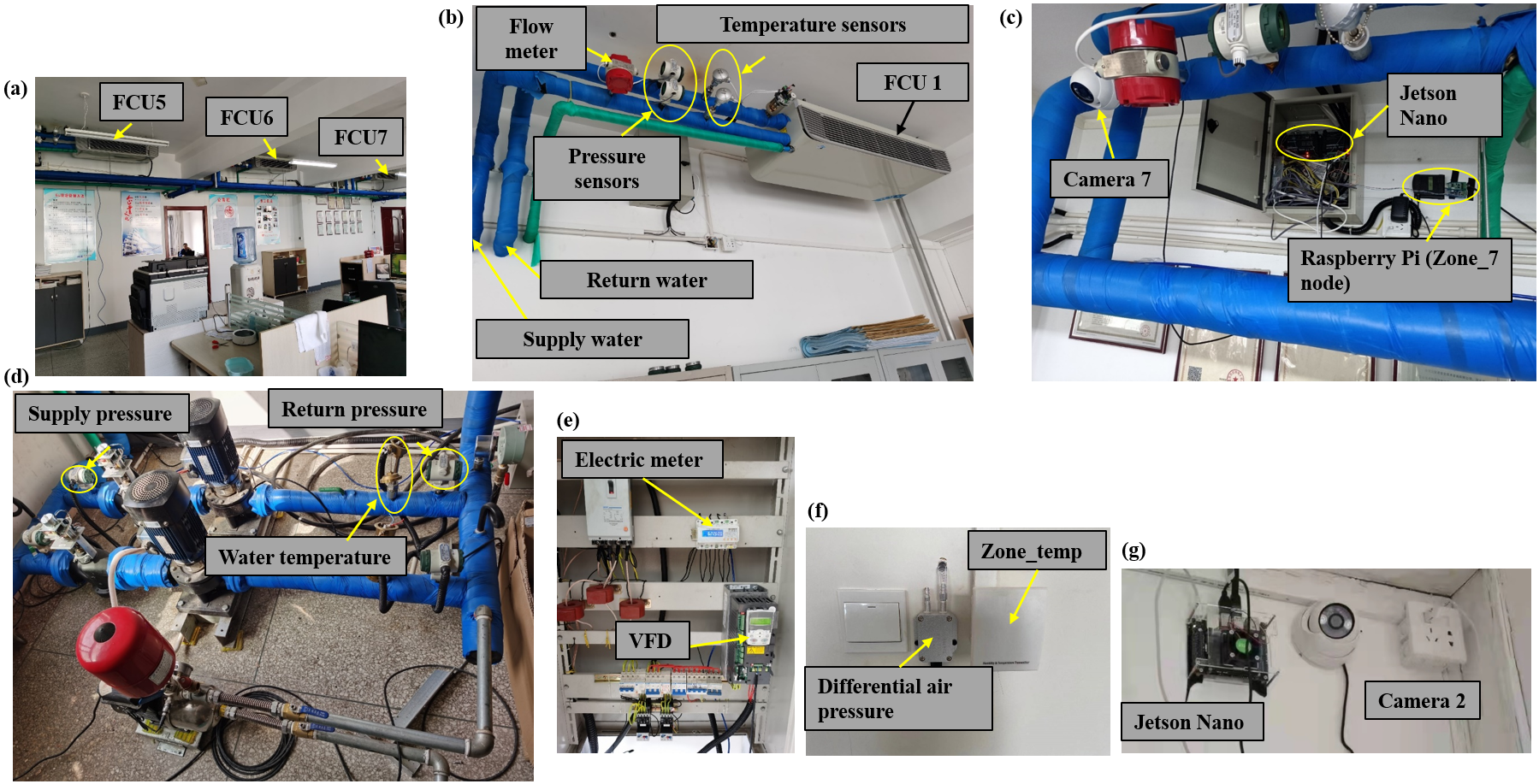}
    \caption{HVAC system configuration. (a) FCUs of zones 5 to 7. (b) Sensors of FCU 1. (c) Video sampling of zone 7. (d) Sensors of water pumps. (e) Electrical cabinet. (f) Temperature sensor. (g) Video camera of zone 2 \cite{zhong2025topology}.}
    \label{fig:real-hvac}
\end{figure*}

As shown in Fig.~\ref{fig:layout}, the building comprises several individual rooms and a hall. One fan coil unit (FCU) is installed in each thermal zone. FCUs~1--4 each serve an individual cellular office, whereas FCUs~5--7 serve zones that allow inter-zone heat exchange. The FCUs receive chilled or heated water from a central refrigeration station comprising a heat pump and a circulating water pump. The pump distributes water to all FCUs through a closed-loop pipe network, while the heat pump provides the thermal source for cooling and heating. An overview of the HVAC system configuration and representative devices is shown in Fig.~\ref{fig:real-hvac}.

The simulator comprises three rigorously coupled components: indoor zone models, parameterized HVAC equipment models, and a dynamic cooling water pipe network~\cite{xing2022honeycomb}. The indoor thermal dynamics are captured using a lumped capacitance approach that accounts for building envelope conduction (via the OTTV method), solar radiation, instantaneous occupancy loads, inter-zone heat transfer, and the sensible cooling/heating capacities delivered by the terminal HVAC units. The hydraulic pipe network utilizes a directed graph topology, solving the complete set of nonlinear hydraulic equations at each time step using the Newton-Raphson method to ensure mass continuity and system pressure-flow balance.

To support simulator calibration and monitor indoor thermal conditions, a comprehensive sensing infrastructure is deployed. At the zone level, each thermal zone is equipped with indoor air temperature sensors (Fig.~\ref{fig:real-hvac}(f)), and occupancy is inferred from ceiling-mounted video cameras (Fig.~\ref{fig:real-hvac}(c, g)). At the terminal side, each FCU is instrumented with a water flow meter, supply and return water temperature sensors, and pressure sensors (Fig.~\ref{fig:real-hvac}(b)). In the central plant, the variable frequency drive (VFD) circulating water pumps (1~Hz control resolution) are equipped with pressure and temperature sensors (Fig.~\ref{fig:real-hvac}(d)), and their electrical power is measured via electrical meters (Fig.~\ref{fig:real-hvac}(e)). All sensors operated continuously during HVAC runtime.

\subsubsection{Observation and Action Spaces}
\label{subsubsec:obs_action}

We formulate office HVAC control as a partially observable Markov
decision process (POMDP). The agent receives observations and issues
control commands every $\Delta t = 5$ minutes.

\textbf{Observation Space.} The observation vector $\mathbf{o}_t$
includes indoor air temperature $T_{\mathrm{in},i}$, occupant count,
and occupant activity states in each of the seven zones, together with
outdoor weather conditions and HVAC operating states. Occupant activities
are classified as sitting, standing, or walking. The simulator uses
these activity categories to calculate occupant-generated internal heat
gains in each zone.

\textbf{Action Space.} The agent independently controls the fan mode
of one fan coil unit (FCU) in each zone. Each FCU has four discrete
modes corresponding to off, low, medium, and high. The resulting
MultiDiscrete action space is $\mathbf{a}_t \in \{0,1,2,3\}^{7}$.

\subsubsection{Performance Metrics and Reward Function}
\label{subsubsec:metrics_reward}

We evaluate controller performance in terms of occupant thermal comfort,
HVAC energy use, and action fluctuation.

\textbf{Thermal Comfort.} We evaluate thermal comfort using the
Predicted Mean Vote (PMV) and Predicted Percentage of Dissatisfied
(PPD) indices~\cite{fanger1970thermal,ashrae2023standard55}. For each
activity $c \in \mathcal{C}
= \{\text{sitting}, \text{standing}, \text{walking}\}$, we use the
corresponding metabolic rate $M_c$ to calculate PMV and its associated
PPD in each zone. We fix the air velocity at $0.15$~m/s, relative
humidity at $40\%$, and clothing insulation at $0.63$~clo. For the
well-insulated office envelope, we approximate the mean radiant
temperature by the zone air temperature.

Let $n_{i,t}^{(c)}$ denote the number of occupants performing activity
$c$ in zone $i$ at time step $t$, and let $\mathrm{PPD}_{i,t}^{(c)}$
be the corresponding PPD. The building-level PPD is the
occupancy-weighted mean across zones and activities:
\begin{equation}
    \mathrm{PPD}_{\mathrm{mean},t} =
    \begin{cases}
        \dfrac{1}{N_t}
        \displaystyle\sum_{i=1}^{J}\sum_{c\in\mathcal{C}}
        n_{i,t}^{(c)}\,\mathrm{PPD}_{i,t}^{(c)}, & N_t > 0, \\[4pt]
        0, & N_t = 0,
    \end{cases}
\end{equation}
where $J=7$ and
$N_t = \sum_{i=1}^{J}\sum_{c\in\mathcal{C}} n_{i,t}^{(c)}$
is the total number of occupants.

\textbf{HVAC Energy Use.} The simulator reports total HVAC electrical
power $P_t$ in kW at each time step. Episode-level energy use is
$E = \sum_t P_t \Delta t$ in kWh, with $\Delta t$ expressed in hours
for this calculation.

\textbf{Reward Function.} The step reward balances thermal discomfort,
HVAC power use, and changes in FCU fan modes:
\begin{equation}
    r_t = -r_{\mathrm{comfort},t}
          -\alpha r_{\mathrm{energy},t}
          -\beta R_{\mathrm{smooth},t},
\end{equation}
where $r_{\mathrm{comfort},t}$ is computed from
$\mathrm{PPD}_{\mathrm{mean},t}$, $r_{\mathrm{energy},t}$ is computed
from $P_t$, and $R_{\mathrm{smooth},t}$ penalizes changes in fan modes
between consecutive control steps. We set $\alpha=20$ and use $\beta=5$
for logarithmic regularization and $\beta=3$ for linear regularization.

\textbf{Action Fluctuation.} We quantify stepwise action fluctuation
as the sum of squared changes in fan-mode indices across the seven FCUs:
\begin{equation}
    AF_t = \sum_{d=1}^{D} \left|a_{t,d} - a_{t-1,d}\right|^2,
    \qquad D=7.
\end{equation}

\begin{table*}[htbp]
  \centering
  \caption{Multi-step prediction performance of world models on SemiBuildingSim and Sinergym. Lower is better for all reported errors.}
  \label{tab:prediction_best_multi_env}
  \begin{tabular*}{\linewidth}{@{\extracolsep{\fill}}l|l|ccccc@{}}
    \toprule
    \multicolumn{2}{@{}l|}{Metric}
    & MLP & LSTM & Transformer & Mamba & \textbf{IEWM (Ours)} \\
    \midrule
    \multicolumn{7}{c}{\textbf{Environment 1: SemiBuildingSim}} \\
    \midrule
    \multirow{2}{*}{%
      \begin{tabular}[c]{@{}l@{}}
        \textit{Overall} \\ \textit{Performance}
      \end{tabular}}
    & MAE ($\downarrow$)
    & 0.046 & 0.058 & 0.045 & 0.038 & \textbf{0.036} \\
    & RMSE ($\downarrow$)
    & 0.090 & 0.108 & 0.088 & 0.080 & \textbf{0.078} \\
    \midrule
    \multirow{3}{*}{%
      \begin{tabular}[c]{@{}l@{}}
        \textit{CVRMSE per} \\ \textit{Dimension (\%) $\downarrow$}
      \end{tabular}}
    & Outdoor Temp
    & 1.397 & 1.267 & 1.386 & 1.334 & \textbf{1.099} \\
    & Return Temp
    & 10.584 & 11.026 & 9.988 & 9.175 & \textbf{9.029} \\
    & Room Temp
    & 5.972 & 8.678 & 5.893 & 5.538 & \textbf{5.479} \\
    \midrule
    \multicolumn{7}{c}{\textbf{Environment 2: Sinergym}} \\
    \midrule
    \multirow{2}{*}{%
      \begin{tabular}[c]{@{}l@{}}
        \textit{Overall} \\ \textit{Performance}
      \end{tabular}}
    & MAE ($\downarrow$)
    & 0.058 & 0.062 & 0.037 & 0.030 & \textbf{0.019} \\
    & RMSE ($\downarrow$)
    & 0.102 & 0.114 & 0.073 & 0.065 & \textbf{0.041} \\
    \multirow{3}{*}{%
      \begin{tabular}[c]{@{}l@{}}
        \textit{CVRMSE per} \\ \textit{Dimension (\%) $\downarrow$}
      \end{tabular}}
    & Outdoor Temp
    & 6.366 & 6.166 & 4.310 & 4.979 & \textbf{3.380} \\
    & Zone Temp
    & 4.920 & 4.500 & 2.632 & 2.671 & \textbf{1.213} \\
    & Humidity
    & 14.289 & 15.790 & 11.531 & 8.919 & \textbf{7.963} \\
    \bottomrule
  \end{tabular*}
\end{table*}

\begin{figure*}[htbp]
    \centering
    \includegraphics[width=1.0\linewidth]{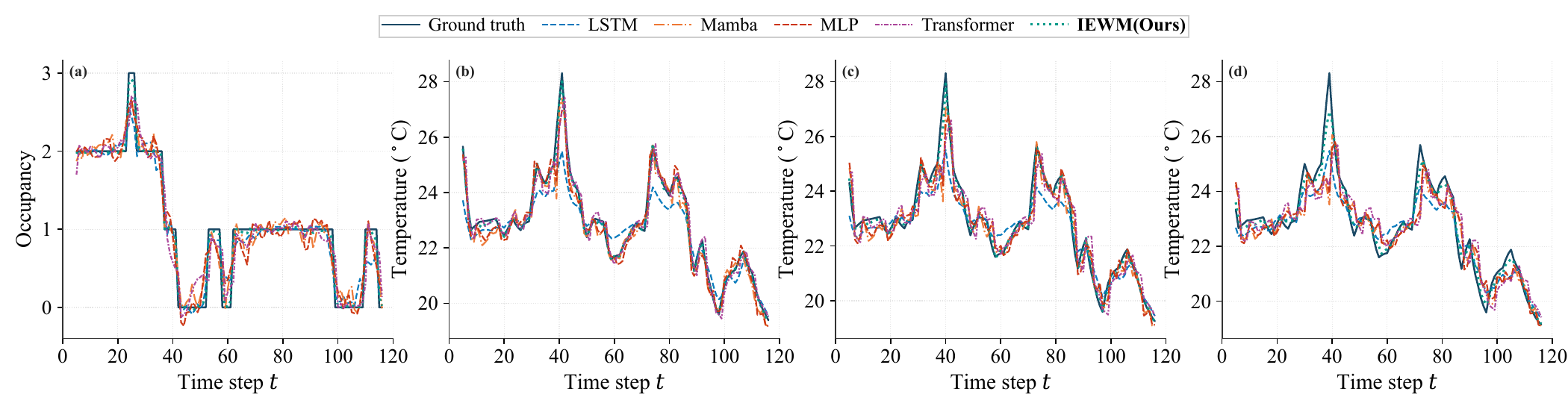}
    \caption{Qualitative multi-step prediction on a representative SemiBuildingSim episode. From left to right: occupancy at $t+1$ and room temperature at $t+1$, $t+2$, and $t+3$ for Room 5. The proposed IEWM remains closest to the ground truth as the prediction horizon increases.}
    \label{fig:rq1_prediction}
\end{figure*}

\subsection{Sinergym 2ZoneDataCenter Environment}
\label{subsec:sinergym_env}

\subsubsection{Environment Construction}
To benchmark our proposed control strategy against established baselines, we utilize the \texttt{2ZoneDataCenter} environment provided by the open-source Sinergym framework~\cite{jimenez2021sinergym}. Sinergym serves as a Python-based virtual testbed that integrates deep reinforcement learning algorithms with the high-fidelity EnergyPlus building simulation engine. 

The selected building model represents a single-story data center with a total surface area of 491.3~$\text{m}^2$. The facility is divided into two asymmetrical thermal zones: the West Zone and the East Zone. Unlike typical commercial buildings, the primary internal heat load in this environment is generated continuously by the hosted server racks (ITE objects), and the building has no windows or internal mass. To manage this intense thermal output, each zone is equipped with an HVAC system consisting of an air economizer, direct and indirect evaporative coolers, a single-speed direct expansion (DX) cooling coil, a chilled water coil, and a Variable Air Volume (VAV) system without reheat air terminal units. 

\subsubsection{Observation and Action Spaces}
The environment is formulated as a discrete-time Markov Decision Process (MDP), where the agent interacts with the EnergyPlus backend via Sinergym's Gymnasium interface.

\textbf{Observation Space:} The state vector $\mathbf{s}_t$ encompasses both exogenous disturbances and the internal thermodynamic state of the data center. Key observations include outdoor weather variables (e.g., outdoor air temperature, relative humidity, solar radiation), the current indoor air temperatures of the West and East zones, and the instantaneous power consumption of the HVAC components and IT equipment.

\textbf{Action Space:} At each control step, the agent selects a discrete
action from Sinergym's predefined action-to-setpoint mapping. Each action
specifies one pair of heating and cooling setpoints shared by the West and
East zones.

\subsubsection{Reward Function and Temperature Penalty}

The data-center controller aims to reduce HVAC electricity use while
maintaining both server zones within their operating temperature band.
Following Sinergym's linear reward formulation, we penalize HVAC power
demand and temperature deviations from the band
$[T_{\text{low}}, T_{\text{high}}] = [20^{\circ}\text{C}, 26^{\circ}\text{C}]$.
For each zone $i \in \{\text{West}, \text{East}\}$, the temperature
deviation at time step $t$ is
\begin{equation}
    \Delta T_{i,t}
    = \max(0, T_{i,t} - T_{\text{high}})
    + \max(0, T_{\text{low}} - T_{i,t}),
\end{equation}
where $T_{i,t}$ is the indoor air temperature of zone $i$.
The deviation is zero within the operating band and increases linearly
with the distance from the nearest bound outside the band.

The step reward combines normalized HVAC power demand with the
normalized sum of temperature deviations across the two zones:
\begin{equation}
    r_t
    = -\alpha \omega \frac{E_t}{E_{\text{max}}}
      -\beta (1-\omega)
      \frac{\sum_{i \in \{\text{West}, \text{East}\}} \Delta T_{i,t}}
           {T_{\text{max}}},
\end{equation}
where $E_t$ is the total HVAC electrical power demand at time step $t$,
and $E_{\text{max}}$ and $T_{\text{max}}$ are normalization constants.
The coefficients $\alpha$ and $\beta$ scale the power and temperature
penalties, respectively, while $\omega \in [0,1]$ sets their relative
weight. We set $\omega = 0.5$, $\alpha = 1$, and
$\beta = 5 \times 10^{-3}$.

\section{Results}
\label{sec:results}
The overall research question in Section 1 is answered through four sub-questions (SQs) that follow the methodological route in Section~\ref{sec:approach}.

\begin{itemize}[leftmargin=*]
    \item \textbf{SQ1:} Can the proposed IEWM accurately predict the delayed thermal baseline under different regions?
    \item \textbf{SQ2:} Can the proposed CAL improve delayed credit assignment and thereby achieve a better occupant-centric trade-off between comfort and energy consumption?
    \item \textbf{SQ3:} How effectively does the proposed smoothness regularization reduce high-frequency actuator fluctuation, and how does it affect energy consumption and occupant comfort?
    \item \textbf{SQ4:} Can the proposed framework maintain occupant-centric control performance when the prediction horizon is extended to cover the long-horizon thermal responses of buildings?
\end{itemize}
Sections~\ref{subsec:sq1}--\ref{subsec:sq4} report the results for SQ1--SQ4, and Section~\ref{subsec:ablation} presents the ablation study.

\begin{table*}[htbp]
  \centering
  \caption{Performance comparisons on SemiBuildingSim (mean $\pm$ std over three seeds). PMV and PPD are computed over occupied periods. Lower is better for all metrics.}
  \label{tab:performance_comparison_wm}
  \begin{tabular*}{\linewidth}{@{\extracolsep{\fill}}lcccc@{}}
    \toprule
    Algorithm & Energy (kWh) $\downarrow$ & Abs PMV $\downarrow$
    & PPD (\%) $\downarrow$ & Action Fluctuation $\downarrow$ \\
    \midrule
    MPC & $156.82 \pm 1.59$ & $0.50 \pm 0.03$ & $13.51 \pm 1.01$ & $10.37 \pm 0.93$ \\
    \midrule
    A2C & $157.81 \pm 7.59$ & $0.50 \pm 0.06$ & $14.05 \pm 1.56$ & $12.89 \pm 3.35$ \\
    PPO & $154.83 \pm 3.10$ & $0.41 \pm 0.01$ & $12.37 \pm 0.33$ & $9.75 \pm 0.85$ \\
    DQN & $155.77 \pm 2.03$ & $0.49 \pm 0.06$ & $13.19 \pm 1.88$ & $9.86 \pm 1.52$ \\
    BDQ & $154.57 \pm 5.23$ & $0.37 \pm 0.01$ & $10.88 \pm 0.78$ & $10.20 \pm 1.17$ \\
    TransformerRL & $149.34 \pm 2.93$ & $0.37 \pm 0.03$ & $10.05 \pm 0.50$ & $9.42 \pm 1.05$ \\
    \midrule
    MBVE & $150.73 \pm 5.88$ & $0.36 \pm 0.01$ & $10.27 \pm 0.14$ & $9.61 \pm 1.86$ \\
    MBPO & $150.99 \pm 3.12$ & $0.38 \pm 0.04$ & $10.47 \pm 0.49$ & $12.47 \pm 1.85$ \\
    DreamerV3 & $149.43 \pm 1.02$ & $0.36 \pm 0.01$ & $9.88 \pm 1.53$ & $8.37 \pm 0.15$ \\
    \midrule
    \textbf{CAL (Ours)}
    & $\mathbf{138.03 \pm 2.99}$
    & $\mathbf{0.26 \pm 0.01}$
    & $\mathbf{7.08 \pm 0.12}$
    & $\mathbf{6.86 \pm 0.80}$ \\
    \bottomrule
  \end{tabular*}
\end{table*}

\begin{figure*}[htbp]
    \centering
    \begin{subfigure}[b]{0.38\textwidth}
        \centering
        \includegraphics[width=\linewidth]{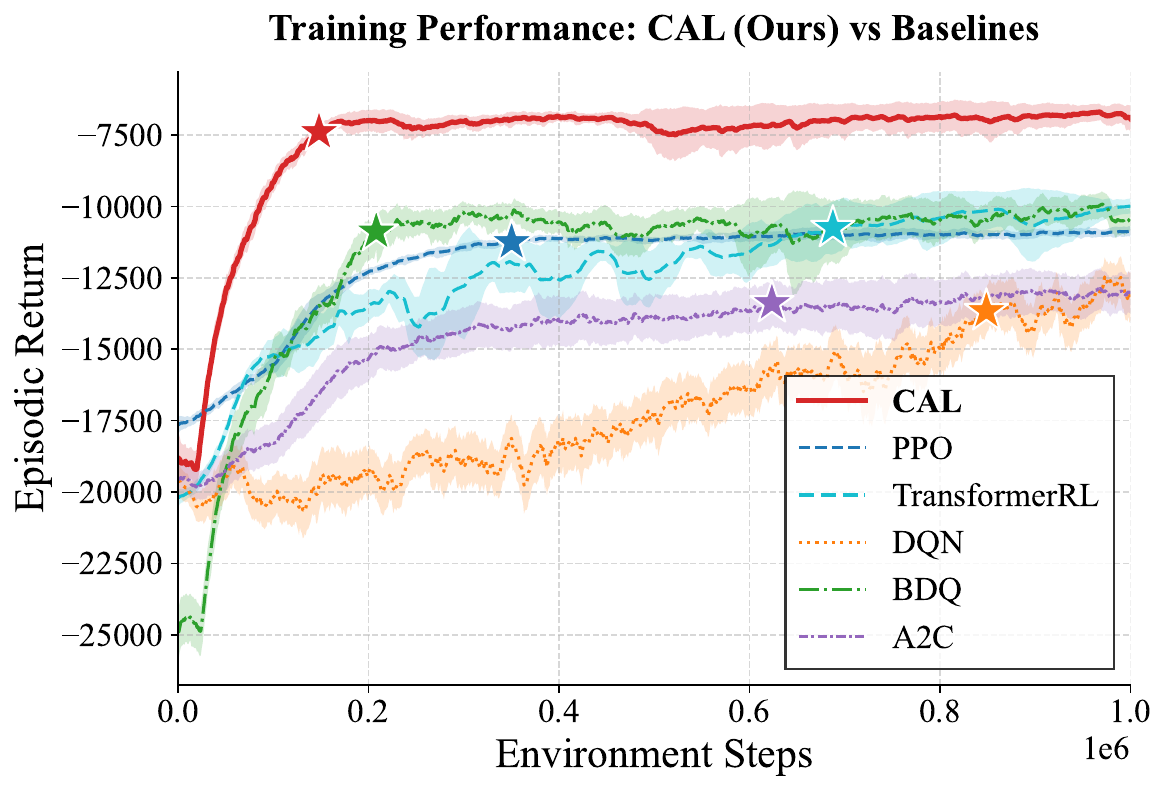}
        \caption{Training return.}
        \label{fig:rq2_sample_efficiency}
    \end{subfigure}
    \begin{subfigure}[b]{0.60\textwidth}
        \centering
        \includegraphics[width=\linewidth]{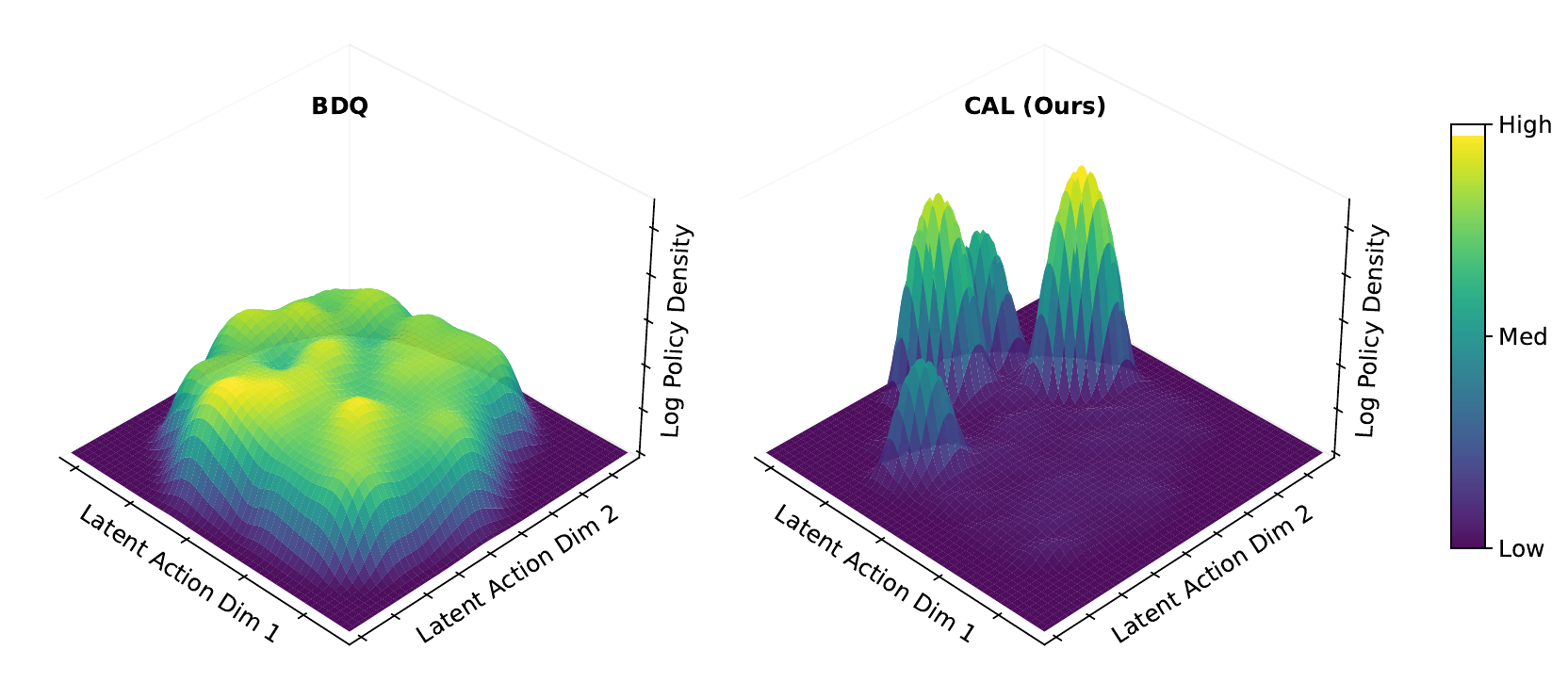}
        \caption{Action-score concentration.}
        \label{fig:rq2_action_concentration}
    \end{subfigure}
    \caption{Mechanistic analysis on SemiBuildingSim. (a) Smoothed training return averaged over 3 seeds; stars denote the first point at which each method reaches $95\%$ of its peak return. 
    (b) Action-score landscape under a specific initial trajectory. Candidate actions are projected to two dimensions with UMAP and weighted by normalized action scores. Compared with BDQ, CAL produces a more concentrated high-value region, highlighting the effectiveness of counterfactual advantage.}
    \label{fig:rq2_mechanism}
\end{figure*}

\subsection{SQ1: Prediction Accuracy of IEWM}
\label{subsec:sq1}
To answer SQ1, we evaluated the proposed IEWM $f_\theta$ on both SemiBuildingSim and Sinergym for predicting the delayed thermal baseline $\hat{\tau_f}$.
As shown in Tab.~\ref{tab:prediction_best_multi_env}, our model outperforms MLP, LSTM~\cite{kim2019real}, Transformer~\cite{sun2023building}, and Mamba~\cite{gu2023mamba}. 

On the SemiBuildingSim environment, IEWM achieves the lowest MAE and RMSE, as well as the lowest CVRMSE across key temperature and humidity dimensions. 
This advantage is important because the quality of the predicted thermal baseline $\hat{\tau_f}$ directly affects downstream action evaluation. 
Compared with MLP, the proposed model better captures temporal dependencies; compared with LSTM, it more reliably preserves delayed thermal response over the history window; and compared with Transformer and Mamba, it maintains multi-step stability across extended prediction horizons. Fig.~\ref{fig:rq1_prediction} demonstrates that while baselines suffer from cumulative multi-step drift, the IEWM tightly tracks the true delayed trajectory and precisely anticipates occupancy dynamics. Furthermore, Sinergym evaluations under Chicago weather confirm its performance on different climate regions.


\begin{figure*}[htbp]
    \centering
    \includegraphics[width=0.8\linewidth]{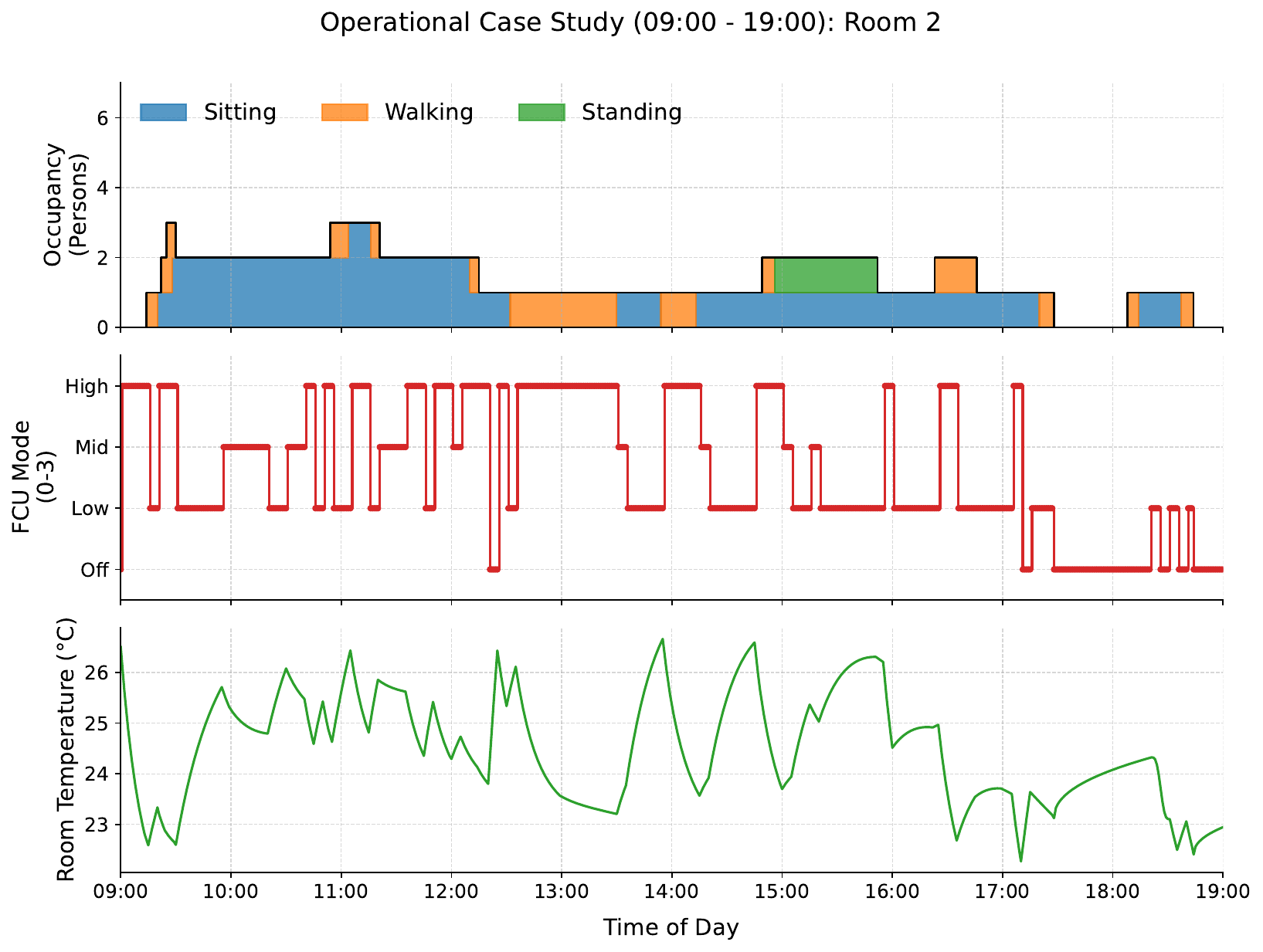}
    \caption{Single-zone control profiles over a representative day, showing occupancy by activity, FCU fan-speed mode, and indoor air temperature. CAL initiates cooling before occupant arrival, adjusts cooling intensity around activity transitions, and switches off the FCU during vacant periods. Before the transition from sitting to walking around 12:30, CAL increases cooling to accommodate the higher metabolic heat associated with walking, illustrating anticipatory, occupant-centric control.}
    \label{fig:rq2_micro_view}
\end{figure*}

\begin{figure*}[htbp]
    \centering
    \includegraphics[width=0.95\linewidth]{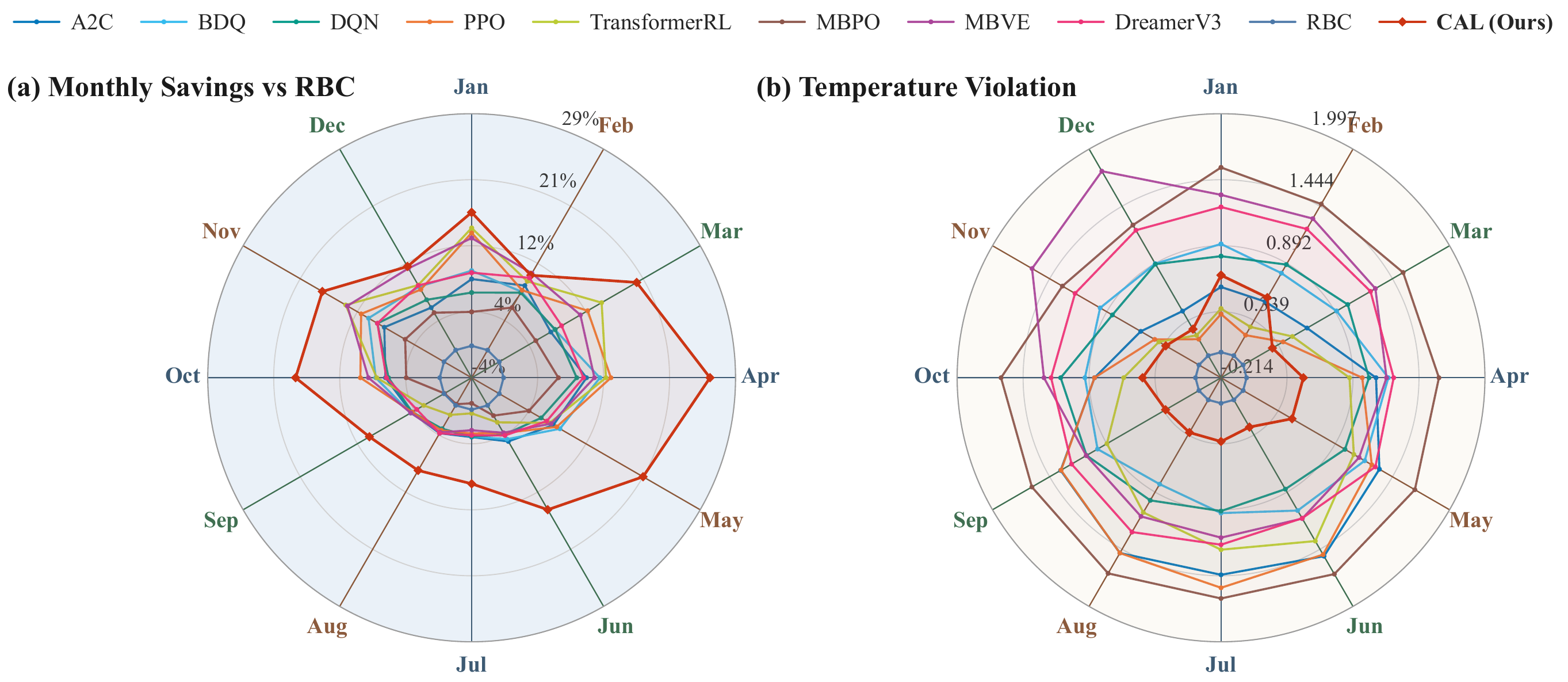}
    \caption{Performance comparison of monthly energy savings relative to the rule-based controller and temperature violation on Sinergym \texttt{2ZoneDataCenterHVAC} under Chicago weather. CAL maintains positive savings across the year and shows the largest advantage during the transition seasons (Apr--Jun and Oct--Nov), while achieving least temperature violation.Radar chart comparing the monthly mean total temperature violation of CAL against multiple baselines in Sinergym (angular axis: Jan–Dec; radial axis: mean violation over timesteps in that month)}
    \label{fig:rq2_monthly_savings}
\end{figure*}
\subsection{SQ2: Control Performance of CAL}
\label{subsec:sq2}
To answer SQ2, we evaluate CAL on SemiBuildingSim against MPC (using the same IEWM with CAL)~\cite{hansen2024tdmpc2}, model-free RL (A2C~\cite{mnih2016asynchronous}, DQN~\cite{mnih2015human}, PPO~\cite{schulman2017proximal}, BDQ~\cite{tavakoli2018action}), history-aware RL (TransformerRL~\cite{ni2023transformers}), and model-based RL (MBVE~\cite{feinberg2018model}, MBPO~\cite{janner2019when}, DreamerV3~\cite{hafner2025mastering}). In Tab.~\ref{tab:performance_comparison_wm}, CAL achieves the best overall trade-off between occupant comfort, energy use, and action smoothness.

Standard model-free RL relies mainly on one-step temporal difference learning, forcing the network to infer future thermal drift only implicitly from past observations, which often leads to reactive control and redundant actuation. 
While advanced sequence models improve feed-forward baselines by encoding historical context into hidden states, they still infer delayed thermal dynamics implicitly and fail to explicitly forecast occupancy dynamics. 
Model-based methods evaluate actions by rolling out imagined trajectories. However, HVAC systems exhibit profound thermal delays, long horizon planning in a combinatorial multi-actuator space inevitably suffers from cumulative prediction errors.

In contrast, our method uses IEWM to predict a delayed thermal baseline $\hat{\tau_f}$ that explicitly captures the building's latent thermal inertia.
As Fig.~\ref{fig:rq2_mechanism}(a) demonstrates, evaluating actions as counterfactual advantages relative to the thermal baseline and propagating delayed feedback via Held-Action TD-$\lambda$, improves sample efficiency, converging faster and achieving a higher return.

Fig.~\ref{fig:rq2_mechanism}(b) further explains this improvement at the action-value level. Under the same state, vanilla BDQ assigns high scores to a broad range of actions, whereas CAL concentrates these scores on a compact subset by distinguishing actions that have marginal benefits over the held-action thermal baseline from redundant alternatives.
Compared with BDQ, CAL reduces action-distribution entropy from $5.39$ to $0.55$ and the effective action count from $219.24$ to $1.74$, while increasing the maximum normalized action weight from $21.88\%$ to $76.29\%$.
These concentration measures suggest that counterfactual adaptive learning reduces redundant exploration, consistent with the improved sample efficiency.

To further illustrate occupant-centric control, we analyze the daily profile of one representative room in SemiBuildingSim (Fig.~\ref{fig:rq2_micro_view}).
CAL adjusts fan modes around occupancy transitions while avoiding unnecessary actuation during vacant periods.
Before occupants arrive, the controller initiates cooling to bring the room temperature towards a comfortable level. 
Around 09:30, as occupants transition from walking to sitting, the fan switches to a lower mode and the room temperature gradually rises, accommodating the reduced metabolic heat production. 
During the vacant period around 17:30--18:15, the FCU remains off to save energy.
When occupancy resumes, brief operation at the lowest fan setting provides sufficient cooling, consistent with the lower cooling demand towards the evening. 
These behaviors illustrate how CAL adapts cooling intensity to occupancy, activity levels and changing thermal conditions.


Experiments on Sinergym (Fig.~\ref{fig:rq2_monthly_savings}) show that CAL achieves consistent year-round energy savings, especially during the transitional seasons (Apr--Jun and Oct--Nov), saving approximately 10\% more energy than the strongest baseline. 
During these months, daily variations in outdoor temperature and solar heat gain lead to moderate thermal loads that fluctuate substantially throughout the day.
While reactive controllers tend to overcorrect under these fluctuating conditions, CAL uses the thermal baseline to suppress redundant heating and cooling cycles while maintaining indoor temperatures within the desired comfort range.
The monthly results further show that CAL achieves the lowest temperature violation and the best trade-off between energy consumption and temperature violation, with consistent performance under different climate regions.

\begin{figure*}[htbp]
    \centering
    \includegraphics[width=\linewidth]{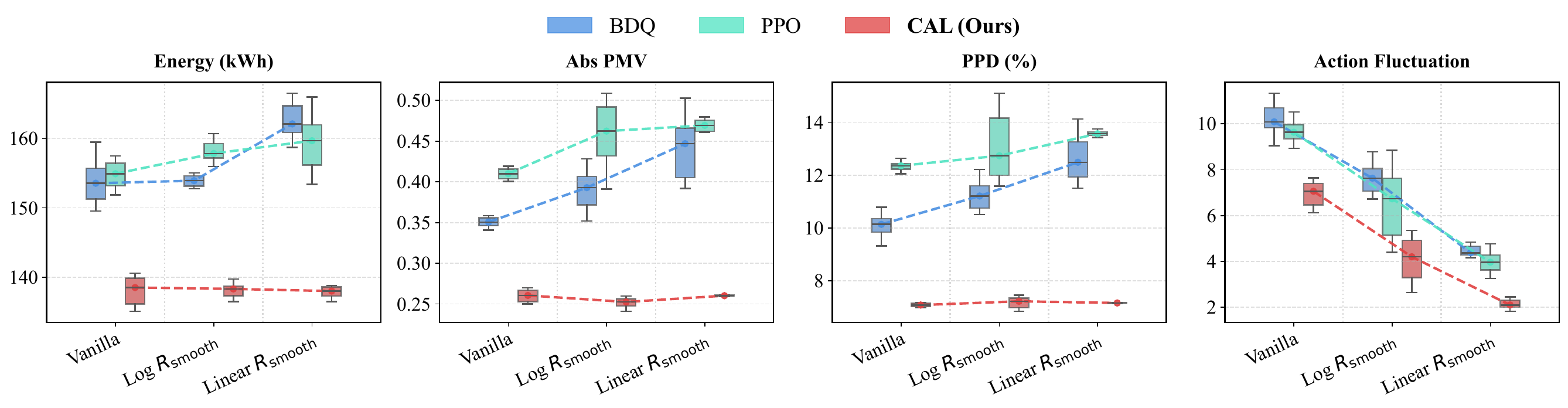}
    \caption{Multi-objective trade-off analysis on SemiBuildingSim. Vanilla denotes training without the smoothness reward. The figure compares CAL with representative baselines under vanilla, logarithmic, and linear smoothness regularization. Regularization consistently reduces fluctuations, with CAL utilizing the linear  smoothness reward achieving the greatest trade-off across all objectives.}
    \label{fig:tradeoff}
\end{figure*}

\subsection{SQ3: Effects of Physics-Aligned Smoothness Regularization}
\label{subsec:sq3}

To anwser SQ3, we 
 evaluate CAL and baselines on SemiBuildingSim under \emph{Vanilla} (no smoothness reward), logarithmic, and linear regularization, as Fig.~\ref{fig:tradeoff} shown. 
Tab.~\ref{tab:performance_comparison_wm_sm1} and Tab.~\ref{tab:performance_comparison_wm_linear} report the full quantitative comparisons.
Both regularization reduce action fluctuation relative to the vanilla setting, but the linear penalty performs the strongest overall trade-off, achieving lower action fluctuation while mantaining the best energy and comfort performance.

In addition to reducing frequent actuator switching, this regularization saves energy by limiting unnecessary control changes by reducing temperature overshoot.
Smoother control helps align actuator operation with the building's thermal inertia. By predicting thermal drift through the baseline $\hat{\tau_f}$, CAL avoids overreacting to short-term deviations, reducing action fluctuations while maintaining its energy and comfort benefits.


\begin{table*}[htbp]
  \centering
  \caption{Performance comparisons on SemiBuildingSim under logarithmic smoothness regularization (mean $\pm$ std over three seeds). PMV and PPD are computed over occupied periods. Lower is better for all metrics.}
  \label{tab:performance_comparison_wm_sm1}
  \begin{tabular*}{\linewidth}{@{\extracolsep{\fill}}lcccc@{}}
    \toprule
    Algorithm & Energy (kWh) $\downarrow$ & Abs PMV $\downarrow$
    & PPD (\%) $\downarrow$ & Action Fluctuation $\downarrow$ \\
    \midrule
    MPC & $167.68 \pm 3.69$ & $0.50 \pm 0.06$ & $13.51 \pm 2.33$ & $7.99 \pm 0.77$ \\
    \midrule
    A2C & $164.04 \pm 7.18$ & $0.51 \pm 0.01$ & $14.08 \pm 0.55$ & $6.61 \pm 0.84$ \\
    PPO & $158.39 \pm 2.45$ & $0.45 \pm 0.06$ & $13.21 \pm 1.94$ & $6.63 \pm 2.28$ \\
    DQN & $162.21 \pm 8.81$ & $0.55 \pm 0.05$ & $15.02 \pm 1.95$ & $8.24 \pm 3.02$ \\
    BDQ & $153.91 \pm 1.22$ & $0.39 \pm 0.04$ & $11.34 \pm 0.91$ & $7.74 \pm 1.05$ \\
    TransformerRL & $151.81 \pm 2.69$ & $0.41 \pm 0.02$ & $11.15 \pm 0.58$ & $6.75 \pm 0.59$ \\
    \midrule
    MBVE & $151.64 \pm 4.05$ & $0.42 \pm 0.02$ & $11.57 \pm 0.62$ & $6.45 \pm 0.56$ \\
    MBPO & $152.31 \pm 4.47$ & $0.43 \pm 0.05$ & $11.08 \pm 0.46$ & $6.61 \pm 1.95$ \\
    DreamerV3 & $150.26 \pm 4.84$ & $0.41 \pm 0.05$ & $10.88 \pm 1.07$ & $6.14 \pm 0.13$ \\
    \midrule
    \textbf{CAL (Ours)}
    & $\mathbf{138.11 \pm 1.71}$
    & $\mathbf{0.25 \pm 0.01}$
    & $\mathbf{7.16 \pm 0.34}$
    & $\mathbf{4.01 \pm 1.43}$ \\
    \bottomrule
  \end{tabular*}
\end{table*}

\begin{table*}[htbp]
  \centering
  \caption{Performance comparisons on SemiBuildingSim under linear smoothness regularization (mean $\pm$ std over three seeds). PMV and PPD are computed over occupied periods. Lower is better for all metrics.}
  \label{tab:performance_comparison_wm_linear}
  \begin{tabular*}{\linewidth}{@{\extracolsep{\fill}}lcccc@{}}
    \toprule
    Algorithm & Energy (kWh) $\downarrow$ & Abs PMV $\downarrow$
    & PPD (\%) $\downarrow$ & Action Fluctuation $\downarrow$ \\
    \midrule
    MPC & $171.08 \pm 3.66$ & $0.58 \pm 0.12$ & $16.01 \pm 3.45$ & $4.37 \pm 0.86$ \\
    \midrule
    A2C & $162.93 \pm 10.31$ & $0.47 \pm 0.06$ & $13.97 \pm 2.44$ & $5.25 \pm 1.68$ \\
    PPO & $159.68 \pm 6.49$ & $0.47 \pm 0.01$ & $13.58 \pm 0.17$ & $4.04 \pm 0.83$ \\
    DQN & $164.60 \pm 15.36$ & $0.56 \pm 0.01$ & $14.98 \pm 0.13$ & $4.99 \pm 2.31$ \\
    BDQ & $162.67 \pm 3.98$ & $0.45 \pm 0.06$ & $12.88 \pm 1.48$ & $4.54 \pm 0.38$ \\
    TransformerRL & $158.83 \pm 6.54$ & $0.47 \pm 0.01$ & $13.02 \pm 0.33$ & $4.52 \pm 0.58$ \\
    \midrule
    MBVE & $156.39 \pm 2.75$ & $0.47 \pm 0.04$ & $13.13 \pm 0.72$ & $4.32 \pm 1.02$ \\
    MBPO & $155.69 \pm 7.11$ & $0.47 \pm 0.06$ & $12.94 \pm 0.89$ & $4.18 \pm 1.05$ \\
    DreamerV3 & $154.96 \pm 1.02$ & $0.44 \pm 0.04$ & $12.83 \pm 0.81$ & $4.17 \pm 0.64$ \\
    \midrule
    \textbf{CAL (Ours)}
    & $\mathbf{137.63 \pm 1.20}$
    & $\mathbf{0.26 \pm 0.01}$
    & $\mathbf{7.16 \pm 0.11}$
    & $\mathbf{2.13 \pm 0.32}$ \\
    \bottomrule
  \end{tabular*}
\end{table*}

\begin{figure*}[htbp]
    \centering
    \includegraphics[width=0.8\linewidth]{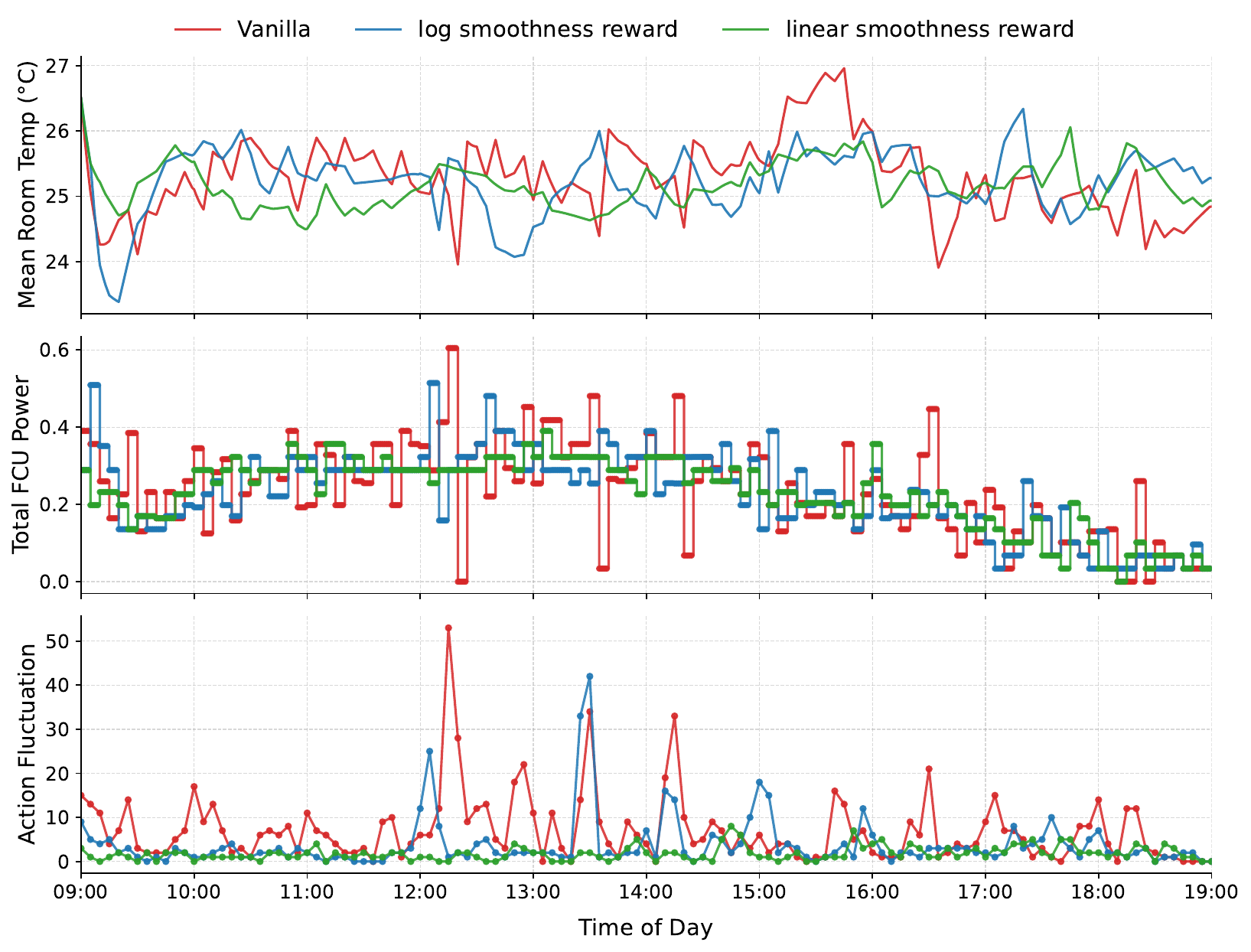}
    \caption{Room-level daily fan-mode and temperature trajectories under different smoothness settings. Vanilla denotes no smoothness reward. Both regularizers suppress high-frequency switching, while the linear penalty yields the most stable actuation pattern.}
    \label{fig:rq3_fanmode}
    \vspace{-1.5em}
\end{figure*}

Fig.~\ref{fig:rq3_fanmode} shows the daily fan-mode profile of a representative room under different smoothness settings. 
From vanilla (red) to logarithmic (blue) and linear (green) regularization, control changes generally become smaller and less frequent, with progressively fewer large action-fluctuation peaks. 
The vanilla setting exhibits abrupt power changes and pronounced temperature fluctuations, particularly around midday and in the afternoon.
Logarithmic regularization moderates these changes, whereas the linear penalty produces longer periods of stable actuation.
Additionally, as smoothness regularization is introduced and strengthened, indoor temperature changes become more gradual while remaining within the comfort range.
These results highlight the benefit of counterfactual learning in distinguishing useful interventions from redundant control changes, enabling smoother actuation while maintaining occupant comfort and energy efficiency.

\subsection{SQ4: Long-Horizon Occupant-Centric Control Performance}
\label{subsec:sq4}
Buildings with high thermal mass, such as those using concrete radiant floors, can exhibit thermal responses over one hour~\cite{zhao2014dynamic}, motivating control over longer horizons. 
To evaluate the effectiveness of our framework under such long-horizon, we increase the control interval to 15 minutes. 
In this specific experiment, we set $H \in \rev{\{}1,2,4,8\rev{\}}$ under the vanilla reward setting on SemiBuildingSim, with these horizons span 15 to 120 minutes. 
Performance remains stable as the horizon increases. 
Energy consumption ranges from $137.37$ to $143.20$ kWh, while PMV and PPD remain within $0.25$–$0.27$ and $6.98\%$–$7.58\%$. 
These results demonstrate that the framework accommodates longer horizons while maintaining energy efficiency, occupant comfort and smooth control.

\begin{table}[htbp]
\centering
\caption{Performance comparison under long-horizon control. In this specific experiment, the control interval increases to 15 minutes. Lower is better.}
\label{tab:horizon_ablation}
\resizebox{\columnwidth}{!}{%
\begin{tabular}{lrrrr}
\hline
Variant & Energy (kWh) & Abs PMV & PPD (\%) & Action Fluctuation \\
\hline
$H=1$ & $143.20 \pm 3.56$ & $0.27 \pm 0.01$ & $7.58 \pm 0.07$ & $5.81 \pm 0.68$ \\
$H=2$ & $139.90 \pm 2.29$ & $0.26 \pm 0.01$ & $7.11 \pm 0.23$ & $5.65 \pm 0.96$ \\
$H=4$ & $137.37 \pm 1.46$ & $0.25 \pm 0.01$ & $6.98 \pm 0.16$ & $5.79 \pm 0.65$ \\
$H=8$ & $138.03 \pm 2.07$ & $0.26 \pm 0.01$ & $7.06 \pm 0.25$ & $4.78 \pm 0.32$ \\
\hline
\end{tabular}%
}
\end{table}

\subsection{Ablation Study}
\label{subsec:ablation}
\textbf{Ablation Study on Core Components:}
We conduct ablation studies under the linear smoothness regularization in Tab.~\ref{tab:ablation_study}.
In removing TD-$\lambda$ ablation, the predicted thermal baseline remains an input to the
action-value network for action selection, while training uses a one-step
TD target in place of the Held-Action TD-$\lambda$ target. Compared with
full CAL, this variant consumes more energy and exhibits greater action
fluctuation, highlighting the benefit of incorporating predicted delayed
thermal effects into the TD target.
Removing the decoupled architecture, the performance drop is sharper on all metrics, showing that separating current state evaluation from baseline-conditioned counterfactual advantage is critical.

\begin{figure}
     \includegraphics[width=\linewidth]{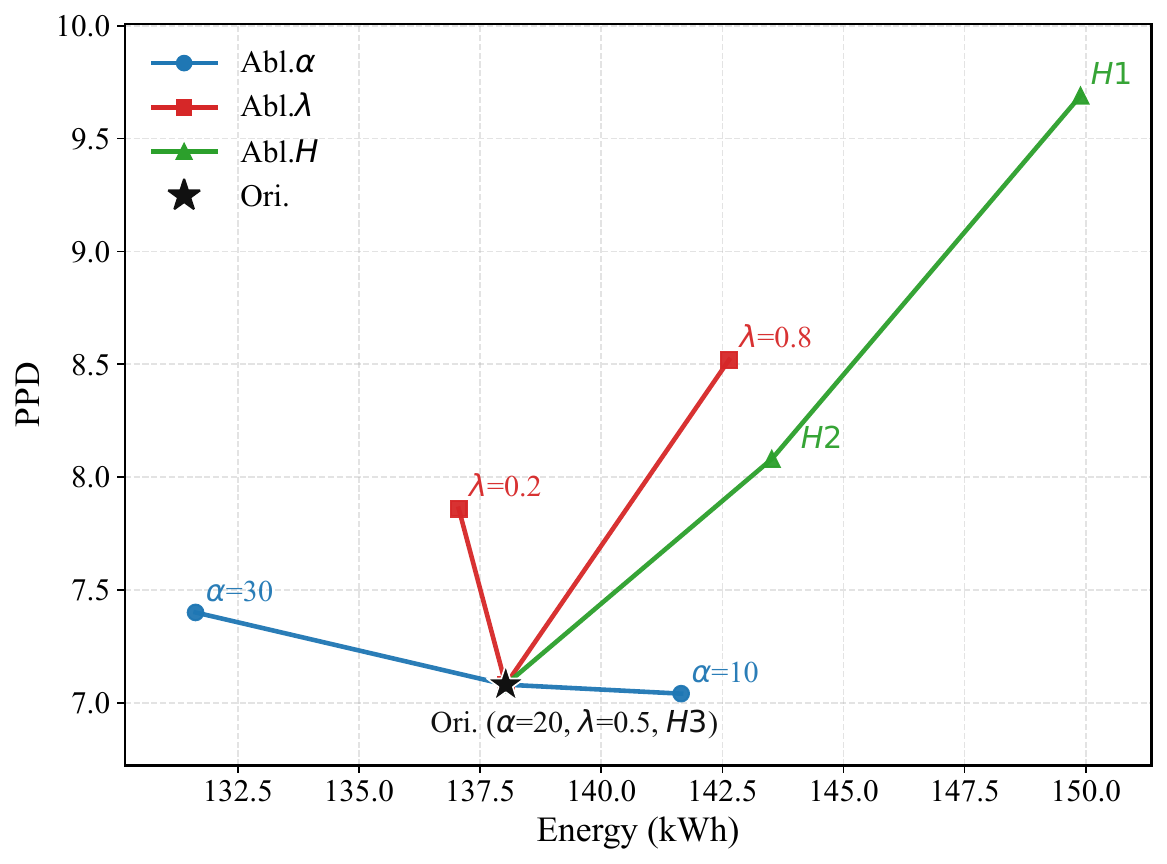}
    \caption{Ablation study on hyper-parameters under linear smoothness  regularization. The control interval is 5 minutes.}
    \vspace{-1em}
    \label{fig:ablation_hparam}
\end{figure}

\textbf{Ablation Study on Hyper-parameters:}
Fig.~\ref{fig:ablation_hparam} shows the energy-comfort trade-off under three hyper-parameters: the energy reward weight $\alpha$, the TD mixing coefficient $\lambda$, and the prediction horizon $H$. The default setting (Ori.) gives the best overall balance.

\begin{table*}[htbp]
  \centering
  \caption{Ablation study under linear smoothness regularization. Lower is better for all metrics.}
  \label{tab:ablation_study}
  \begin{tabular*}{\linewidth}{@{\extracolsep{\fill}}lcccc@{}}
    \toprule
    Variant & Energy (kWh) $\downarrow$ & Abs PMV $\downarrow$
    & PPD (\%) $\downarrow$ & Action Fluctuation $\downarrow$ \\
    \midrule
    w/o Decoupled Architecture
    & $144.56 \pm 3.72$ & $0.33 \pm 0.02$
    & $8.15 \pm 0.22$ & $3.72 \pm 0.99$ \\
    w/o Held-Action TD-$\lambda$
    & $141.25 \pm 4.60$ & $0.28 \pm 0.03$
    & $7.90 \pm 0.34$ & $2.97 \pm 0.41$ \\
    \midrule
    \textbf{CAL (Full)}
    & $\mathbf{137.63 \pm 1.20}$ & $\mathbf{0.26 \pm 0.01}$
    & $\mathbf{7.16 \pm 0.11}$ & $\mathbf{2.13 \pm 0.32}$ \\
    \bottomrule
  \end{tabular*}
  \vspace{-1em}
\end{table*}

\section{Discussion}
\label{sec:discussion}

\subsection{Comparison with previous studies}
\label{subsec:disc_compare}
With the annual global mean temperature in 2024 exceeding 1.5$^\circ$C above pre-industrial levels\rev{~\cite{wmo2025,tollefson2025earth}}, the transition toward low-carbon energy systems has become increasingly urgent. Given that the buildings and construction sector accounts for over one-third of global CO$_2$ emissions\rev{~\cite{2022un}}, optimizing the built environment remains a critical frontier for global climate mitigation. This work advances the United Nations' Sustainable Development Goals (SDGs) 11 and 13 by introducing our method, an AI-for-sustainability framework\rev{~\cite{wang2023occupant,fuso2019connecting}}. By demonstrating  that energy savings can be achieved alongside improvements in occupant comfort, our method helps mitigate the traditional trade-off between energy consumption and occupant comfort.

Methodologically, this study offers new insights into how world models and counterfactual adaptive learning can jointly improve control under partially observed and delayed dynamics.
We train a domain-specific world model to establish a predictive baseline for evaluating control interventions. Its MambaFormer-based indoor environmental world model estimates how the physical environment will evolve with the current action held, providing a short-horizon baseline that reflects latent thermal inertia, and predicts reliable occupancy together with the thermal variables, which are essential for downstream energy-efficient indoor control.

Previous model-free and history-based RL controllers infer this drift implicitly from stacked observations or recurrent and transformer memories~\cite{hausknecht2015deep,ni2023transformers}, whereas our world model predicts it explicitly to inform energy-efficient control decisions. Its multi-step predictions were more accurate than those of the baseline predictors in both environments (Table~\ref{tab:prediction_best_multi_env}).
Counterfactual adaptive learning then uses this prediction as a shared input to the branch-wise advantage streams of BDQ~\cite{tavakoli2018action}, so that all candidate actions are scored against the predicted baseline, with the held action as the counterfactual reference for evaluating energy--comfort trade-offs. Unlike Dreamer, TD-MPC, MBVE and MBPO~\cite{hafner2025mastering,hansen2022temporal,hansen2024tdmpc2,feinberg2018model,janner2019when}, CAL does not roll out candidate action sequences for planning or value estimation, thereby reducing the number of model evaluations and limiting exposure to prediction errors that accumulate over longer rollouts.
Consequently, CAL achieved lower energy use and discomfort than the best baseline on each metric (Table~\ref{tab:performance_comparison_wm}).

\subsection{Broader impacts}
\label{subsec:disc_practice}
The broader insight is that a world model can make ongoing dynamics in delayed thermal systems explicit by predicting them from partial observations and operating histories. CAL uses this prediction as a common input to score candidate actions without optimizing long action sequences online.

Delayed credit assignment and physics-aligned action regularization further support this decision process. 
On SemiBuildingSim, CAL achieved a strong multi-objective trade-off under linear smoothness regularization, reducing energy consumption by 11.2\%, occupant discomfort by 44.2\%, and action fluctuation by 47.3\% relative to the best baseline for each metric. 
Suppressing high-frequency actuator fluctuations through physics-aligned regularization may also reduce mechanical wear.
Scaling this deployment to everyday environments, such as commercial offices and government institutions, could reduce utility costs and financial burdens, translating environmental sustainability into economic savings on a societal scale.

Looking forward, this combination of world model and counterfactual action evaluation provides a transferable computational strategy for other delayed thermal and physical systems with latent dynamics, changing demands, and costly actuation.
Data-center cooling offers a future application: variations in computational workload create changing heat gains, while thermal storage and cooling dynamics delay the effects of control actions. A world model could predict temperature evolution under unchanged cooling settings, while counterfactual adaptive learning could assess whether changes in fan speed or coolant flow provide sufficient thermal benefit to justify their energy and actuation costs. 

In battery thermal management, the same approach could use temperature predictions under current cooling inputs and changing electrical loads to guide cooling adjustments. In industrial process control, predictions under unchanged heating or flow settings could provide a reference for evaluating corrective actions before delayed process responses become fully observable.
Across these domains, the transferable element is the coupling of a learned reference evolution with adaptive evaluation of alternative actions. The predictive model, objectives, and operating constraints would be adapted to each system.

Ultimately, this study offers a transferable computational principle, combining world-model prediction with counterfactual adaptive learning to guide resource-efficient intervention in delayed physical systems.
By sharing our methodology, we aim to support its adaptation across engineering domains and broaden access to sustainable and human-centric control technologies.

\subsection{Limitations}
\label{lim}
The framework performs well in both the calibrated Hebei office simulation
and the Sinergym data center under Chicago weather conditions. Future work
will extend validation to a wider range of buildings and HVAC systems
through field trials. The world model is kept fixed during policy learning
to stabilize training targets, and future work can explore how to retain
this stability while adapting the model to changes in equipment or
operating conditions.

\vspace{-1em}
\section{Conclusions}
\label{sec:conclusion}
We present a framework combining indoor environmental world model and counterfactual adaptive learning for building energy efficiency.
To address delayed thermodynamics and partial observability, our method utilizes the IEWM to predict a thermal baseline, combined with counterfactual advantage estimation via BDQ and a Held-Action TD-$\lambda$ target.
In the calibrated SemiBuildingSim, the IEWM achieved the lowest multi-step prediction errors among the compared sequence models, and CAL under linear smoothness regularization reduced HVAC energy consumption by 11.2\%, occupant discomfort (PPD) by 44.2\% and action fluctuation by 47.3\% relative to the best baseline on each metric. The smoothness regularization lowered the action fluctuation of CAL by 69\% with almost no change in energy use or comfort, and the control performance remained stable as the prediction horizon was extended from 15 to 120~min. In the data center under Chicago weather, CAL saved the most energy relative to the rule-based controller in every month and had the lowest temperature violation, indicating that the framework remains effective across climate regions and building types.
Finally, our method offers a robust, sample-efficient AI approach to reducing building energy use worldwide. This study has the potential to advance the scientific development of AI for sustainability, for achieving SDG 11 and 13.

\clearpage

\bibliographystyle{unsrt}
\bibliography{r1}

@book{lee2023climate,
  title={Climate change 2023: synthesis report. Contribution of working groups I, II and III to the sixth assessment report of the intergovernmental panel on climate change},
  author={Lee, Hoesung and Calvin, Katherine and Dasgupta, Dipak and Krinner, Gerhard and Mukherji, Aditi and Thorne, Peter and Trisos, Christopher and Romero, Jos{\'e} and Aldunce, Paulina and Barrett, Ko and others},
  year={2023}
}

@misc{unfccc2023gst,
  title = {Outcome of the First Global Stocktake},
  author = {{Conference of the Parties serving as the meeting of the Parties to the Paris Agreement (CMA)}},
  year = {2023},
  howpublished = {UNFCCC Decision 1/CMA.5},
  note = {FCCC/PA/CMA/2023/L.17}
}

@inproceedings{sun2026sense,
  title={SENSE: Satellite-based ENergy Synthesis for Sustainable Environment},
  author={Sun, Kailai and He, Mingyi and Huang, Heye and Rong, Can and Prakash, Alok and Guo, Baoshen and Wang, Shenhao and Zhao, Jinhua},
  booktitle={Proceedings of the 32nd ACM SIGKDD Conference on Knowledge Discovery and Data Mining V. 2},
  pages={12030--12041},
  year={2026}
}

@article{zhong2025topology,
  title={Topology-aware hypergraph reinforcement learning for indoor occupant-centric HVAC control},
  author={Zhong, Dianyu and Xing, Tian and Sun, Kailai and Zhang, Ziyou and Zhao, Qianchuan and Kang, Jian},
  journal={Energy and Buildings},
  pages={116219},
  year={2025},
  publisher={Elsevier}
}

@article{xing2022honeycomb,
  title={Honeycomb: An open-source distributed system for smart buildings},
  author={Xing, Tian and Yan, Hu and Sun, Kailai and Wang, Yifan and Wang, Xuetao and Zhao, Qianchuan},
  journal={Patterns},
  volume={3},
  number={11},
  year={2022},
  publisher={Elsevier}
}

@misc{2022un,
author={UN Environment Programme},
title = {“2022 GLOBAL STATUS REPORT FOR BUILDINGS AND CONSTRUCTION”},
year = {2022},
note = {\url{https://globalabc.org}}
}

@article{perez2008hvac,
  title={A review on buildings energy consumption information},
  author={P{\'e}rez-Lombard, Luis and Ortiz, Jos{\'e} and Pout, Christine},
  journal={Energy and buildings},
  volume={40},
  number={3},
  pages={394--398},
  year={2008},
  publisher={Elsevier}
}

@article{wang2023occupant,
  title={An occupant-centric control strategy for indoor thermal comfort, air quality and energy management},
  author={Wang, Zu and Calautit, John and Tien, Paige Wenbin and Wei, Shuangyu and Zhang, Wuxia and Wu, Yupeng and Xia, Liang},
  journal={Energy and Buildings},
  volume={285},
  pages={112899},
  year={2023},
  publisher={Elsevier}
}

@article{EPA_indoor,
  title={Buildings and their impact on the environment: A statistical summary},
  author={EPA, U},
  journal={US Environmental Protection Agency Green Building Workgroup},
  year={2009}
}

@article{fanger1970thermal,
  title={Thermal comfort. Analysis and applications in environmental engineering.},
  author={Fanger, Poul O},
  year={1970}
}

@article{nagy2023ten,
  title={Ten questions concerning occupant-centric control and operations},
  author={Nagy, Zoltan and Gunay, Burak and Miller, Clayton and Hahn, Jakob and Ouf, Mohamed M and Lee, Seungjae and Hobson, Brodie W and Abuimara, Tareq and Bandurski, Karol and Andr{\'e}, Ma{\'\i}ra and others},
  journal={Building and Environment},
  volume={242},
  pages={110518},
  year={2023},
  publisher={Elsevier}
}

@book{ashrae2023standard55,
  title        = {{ANSI/ASHRAE Standard 55-2023: Thermal Environmental Conditions for Human Occupancy}},
  author       = {{American Society of Heating, Refrigerating and Air-Conditioning Engineers}},
  year         = {2023},
  publisher    = {ASHRAE},
  address      = {Atlanta, GA},
  url          = {https://www.ashrae.org/technical-resources/bookstore/standard-55-thermal-environmental-conditions-for-human-occupancy}
}

@article{bayasgalan2024comprehensive,
  title={Comprehensive review of building energy management models: Grid-interactive efficient building perspective},
  author={Bayasgalan, Anujin and Park, Yoo Shin and Koh, Seak Bai and Son, Sung-Yong},
  journal={Energies},
  volume={17},
  number={19},
  pages={4794},
  year={2024},
  publisher={MDPI}
}

@article{drg2020review,
  title={All you need to know about model predictive control for buildings},
  author={Drgo{\v{n}}a, J{\'a}n and Arroyo, Javier and Figueroa, Iago Cupeiro and Blum, David and Arendt, Krzysztof and Kim, Donghun and Oll{\'e}, Enric Perarnau and Oravec, Juraj and Wetter, Michael and Vrabie, Draguna L and others},
  journal={Annual reviews in control},
  volume={50},
  pages={190--232},
  year={2020},
  publisher={Elsevier}
}

@article{wang2021reinforcement,
  title={Reinforcement learning for building controls: The opportunities and challenges},
  author={Wang, Liangxu and Hong, Tianzhen},
  journal={Applied Energy},
  volume={269},
  pages={115036},
  year={2020},
  publisher={Elsevier},
  doi={10.1016/j.apenergy.2020.115036},
}

@article{vazquez2019reinforcement,
  title={Reinforcement learning for demand response: A review of algorithms and modeling techniques},
  author={V{\'a}zquez-Canteli, Jos{\'e} R and Nagy, Zolt{\'a}n},
  journal={Applied energy},
  volume={235},
  pages={1072--1089},
  year={2019},
  publisher={Elsevier}
}

@article{verbeke2018thermal,
  title={Thermal inertia in buildings: A review of impacts across climate and building use},
  author={Verbeke, Stijn and Audenaert, Amaryllis},
  journal={Renewable and sustainable energy reviews},
  volume={82},
  pages={2300--2318},
  year={2018},
  publisher={Elsevier}
}

@article{hanafi2024advancing,
  title={Advancing sustainable energy management: a comprehensive review of artificial intelligence techniques in building},
  author={Hanafi, Ahmed M and Moawed, Mohamed Ahmed and Abdellatif, Osama Ezzat},
  journal={Engineering Research Journal (Shoubra)},
  volume={53},
  number={2},
  pages={26--46},
  year={2024},
  publisher={Benha University, Faculty of Engineering (Shoubra)}
}

@article{nguyen2024modelling,
  title={Modelling building HVAC control strategies using a deep reinforcement learning approach},
  author={Nguyen, Anh Tuan and Pham, Duy Hoang and Oo, Bee Lan and Santamouris, Mattheos and Ahn, Yonghan and Lim, Benson TH},
  journal={Energy and Buildings},
  volume={310},
  pages={114065},
  year={2024},
  publisher={Elsevier}
}

@article{sun2024individual,
  title={Individual room air-conditioning control in high-insulation residential building during winter: A deep reinforcement learning-based control model for reducing energy consumption},
  author={Sun, Luning and Hu, Zehuan and Mae, Masayuki and Imaizumi, Taiji},
  journal={Energy and Buildings},
  volume={323},
  pages={114799},
  year={2024},
  publisher={Elsevier}
}

@article{wang2025safe,
  title={Safe deep reinforcement learning for building energy management},
  author={Wang, Xiangwei and Wang, Peng and Huang, Renke and Zhu, Xiuli and Arroyo, Javier and Li, Ning},
  journal={Applied Energy},
  volume={377},
  pages={124328},
  year={2025},
  publisher={Elsevier}
}

@article{wang2024experimental,
  title={Experimental evaluation of offline reinforcement learning for hvac control in buildings},
  author={Wang, Jun and Li, Linyan and Liu, Qi and Yang, Yu},
  journal={arXiv preprint arXiv:2408.07986},
  year={2024}
}

@article{togashi2025reward,
  title={Reward function design in reinforcement learning for HVAC Control: A review of thermal comfort and energy efficiency Trade-offs},
  author={Togashi, Eisuke},
  journal={Energy and Buildings},
  pages={116439},
  year={2025},
  publisher={Elsevier}
}

@inproceedings{yuan2024review,
  title={A review of current research on occupant-centric control for improving comfort and energy efficiency},
  author={Yuan, Yue and Song, Chengcheng and Gao, Liying and Zeng, Kejun and Chen, Yixing},
  booktitle={Building Simulation},
  volume={17},
  number={10},
  pages={1675--1692},
  year={2024},
  organization={Springer}
}

@inproceedings{liang2025human,
  title={Human-in-the-loop ai for hvac management enhancing comfort and energy efficiency},
  author={Liang, Xinyu and de Nijs, Frits and Say, Buser and Wang, Hao},
  booktitle={Proceedings of the 16th ACM International Conference on Future and Sustainable Energy Systems},
  pages={359--370},
  year={2025}
}

@article{meimand2024personal,
  title={A personal touch to demand response: An occupant-centric control strategy for HVAC systems using personalized comfort models},
  author={Meimand, Mostafa and Jazizadeh, Farrokh},
  journal={Energy and Buildings},
  volume={303},
  pages={113769},
  year={2024},
  publisher={Elsevier}
}

@article{kaack2022aligning,
  title={Aligning artificial intelligence with climate change mitigation},
  author={Kaack, Lynn H and Donti, Priya L and Strubell, Emma and Kamiya, George and Creutzig, Felix and Rolnick, David},
  journal={Nature Climate Change},
  volume={12},
  number={6},
  pages={518--527},
  year={2022},
  publisher={Nature Publishing Group UK London}
}

@article{creutzig2022demand,
  title={Demand-side solutions to climate change mitigation consistent with high levels of well-being},
  author={Creutzig, Felix and Niamir, Leila and Bai, Xuemei and Callaghan, Max and Cullen, Jonathan and D{\'\i}az-Jos{\'e}, Julio and Figueroa, Maria and Grubler, Arnulf and Lamb, William F and Leip, Adrian and others},
  journal={Nature climate change},
  volume={12},
  number={1},
  pages={36--46},
  year={2022},
  publisher={Nature Publishing Group UK London}
}

@article{fuso2019connecting,
  title={Connecting climate action with other Sustainable Development Goals},
  author={Fuso Nerini, Francesco and Sovacool, Benjamin and Hughes, Nick and Cozzi, Laura and Cosgrave, Ellie and Howells, Mark and Tavoni, Massimo and Tomei, Julia and Zerriffi, Hisham and Milligan, Ben},
  journal={Nature sustainability},
  volume={2},
  number={8},
  pages={674--680},
  year={2019},
  publisher={Nature Publishing Group UK London}
}

@article{tollefson2025earth,
  title={Earth breaches 1.5° C climate limit for the first time: what does it mean?},
  author={Tollefson, Jeff},
  journal={Nature},
  volume={637},
  number={8047},
  pages={769--770},
  year={2025},
  publisher={Nature}
}

@misc{wmo2025,
  author       = {{World Meteorological Organization}},
  title        = {WMO confirms 2024 as warmest year on record at about 1.55°C above pre-industrial level},
  year         = {2025},
  url          = {https://wmo.int/news/media-centre/wmo-confirms-2024-warmest-year-record-about-155degc-above-pre-industrial-level}
}

@article{ding2025understanding,
  title={Understanding world or predicting future? a comprehensive survey of world models},
  author={Ding, Jingtao and Zhang, Yunke and Shang, Yu and Zhang, Yuheng and Zong, Zefang and Feng, Jie and Yuan, Yuan and Su, Hongyuan and Li, Nian and Sukiennik, Nicholas and others},
  journal={ACM Computing Surveys},
  volume={58},
  number={3},
  pages={1--38},
  year={2025},
  publisher={ACM New York, NY}
}

@inproceedings{janner2019when,
  title={When to trust your model: Model-based policy optimization},
  author={Janner, Michael and Fu, Justin and Zhang, Marvin and Levine, Sergey},
  booktitle={Advances in neural information processing systems},
  volume={32},
  year={2019}
}

@article{gu2023mamba,
  title={Mamba: Linear-time sequence modeling with selective state spaces},
  author={Gu, Albert and Dao, Tri},
  journal={arXiv preprint arXiv:2312.00752},
  year={2023}
}

@article{buckman2018sample,
  title={Sample-efficient reinforcement learning with stochastic ensemble value expansion},
  author={Buckman, Jacob and Hafner, Danijar and Tucker, George and Brevdo, Eugene and Lee, Honglak},
  journal={Advances in neural information processing systems},
  volume={31},
  year={2018}
}

@inproceedings{van2016deep,
  title={Deep reinforcement learning with double Q-learning},
  author={van Hasselt, Hado and Guez, Arthur and Silver, David},
  booktitle={Proceedings of the AAAI Conference on Artificial Intelligence},
  volume={30},
  number={1},
  year={2016}
}

@article{crawley2001energyplus,
  title={EnergyPlus: creating a new-generation building energy simulation program},
  author={Crawley, Drury B and Lawrie, Linda K and Winkelmann, Frederick C and Buhl, Walter F and Huang, Y Joe and Pedersen, Curtis O and Strand, Richard K and Liesen, Richard J and Fisher, Daniel E and Witte, Michael J and others},
  journal={Energy and buildings},
  volume={33},
  number={4},
  pages={319--331},
  year={2001},
  publisher={Elsevier}
}

@inproceedings{jimenez2021sinergym,
  title={Sinergym: a building simulation and control framework for training reinforcement learning agents},
  author={Jim{\'e}nez-Raboso, Javier and Campoy-Nieves, Alejandro and Manjavacas-Lucas, Antonio and G{\'o}mez-Romero, Juan and Molina-Solana, Miguel},
  booktitle={Proceedings of the 8th ACM international conference on systems for energy-efficient buildings, cities, and transportation},
  pages={319--323},
  year={2021}
}

@article{campoy2025sinergym,
  title={Sinergym--A virtual testbed for building energy optimization with Reinforcement Learning},
  author={Campoy-Nieves, Alejandro and Manjavacas, Antonio and Jim{\'e}nez-Raboso, Javier and Molina-Solana, Miguel and G{\'o}mez-Romero, Juan},
  journal={Energy and Buildings},
  volume={327},
  pages={115075},
  year={2025},
  publisher={Elsevier}
}

@inproceedings{hausknecht2015deep,
  title={Deep Recurrent Q-Learning for Partially Observable MDPs.},
  author={Hausknecht, Matthew J and Stone, Peter},
  booktitle={AAAI fall symposia},
  volume={45},
  pages={141},
  year={2015}
}

@article{sutton1991dyna,
  title={Dyna, an integrated architecture for learning, planning, and reacting},
  author={Sutton, Richard S},
  journal={ACM Sigart Bulletin},
  volume={2},
  number={4},
  pages={160--163},
  year={1991},
  publisher={ACM New York, NY, USA}
}

@article{chua2018deep,
  title={Deep reinforcement learning in a handful of trials using probabilistic dynamics models},
  author={Chua, Kurtland and Calandra, Roberto and McAllister, Rowan and Levine, Sergey},
  journal={Advances in neural information processing systems},
  volume={31},
  year={2018}
}

@inproceedings{kurutach2018model,
  title={Model-Ensemble Trust-Region Policy Optimization},
  author={Kurutach, Thanard and Clavera, Ignasi and Duan, Yan and Tamar, Aviv and Abbeel, Pieter},
  booktitle={International Conference on Learning Representations},
  year={2018}
}

@article{feinberg2018model,
  title={Model-based value estimation for efficient model-free reinforcement learning},
  author={Feinberg, Vladimir and Wan, Alvin and Stoica, Ion and Jordan, Michael I and Gonzalez, Joseph E and Levine, Sergey},
  journal={arXiv preprint arXiv:1803.00101},
  year={2018}
}

@inproceedings{hansen2022temporal,
  title={Temporal Difference Learning for Model Predictive Control},
  author={Hansen, N and Wang, X and Su, H},
  booktitle={International Conference on Machine Learning, PMLR},
  year={2022}
}

@article{hafner2023mastering,
  title={Mastering diverse domains through world models},
  author={Hafner, Danijar and Pasukonis, Jurgis and Ba, Jimmy and Lillicrap, Timothy},
  journal={arXiv preprint arXiv:2301.04104},
  year={2023}
}

@article{lecun2022path,
  title={A path towards autonomous machine intelligence version 0.9. 2, 2022-06-27},
  author={LeCun, Yann and others},
  journal={Open Review},
  volume={62},
  number={1},
  pages={1--62},
  year={2022}
}

@inproceedings{hansen2024tdmpc2,
  title={TD-MPC2: Scalable, Robust World Models for Continuous Control},
  author={Hansen, Nicklas and Su, Hao and Wang, Xiaolong},
  booktitle={The Twelfth International Conference on Learning Representations}
}

@inproceedings{mnih2016asynchronous,
  title={Asynchronous methods for deep reinforcement learning},
  author={Mnih, Volodymyr and Badia, Adria Puigdomenech and Mirza, Mehdi and Graves, Alex and Lillicrap, Timothy and Harley, Tim and Silver, David and Kavukcuoglu, Koray},
  booktitle={International conference on machine learning},
  pages={1928--1937},
  year={2016},
  organization={PmLR}
}

@article{mnih2015human,
  title={Human-level control through deep reinforcement learning},
  author={Mnih, Volodymyr and Kavukcuoglu, Koray and Silver, David and Rusu, Andrei A and Veness, Joel and Bellemare, Marc G and Graves, Alex and Riedmiller, Martin and Fidjeland, Andreas K and Ostrovski, Georg and others},
  journal={nature},
  volume={518},
  number={7540},
  pages={529--533},
  year={2015},
  publisher={Nature Publishing Group}
}

@inproceedings{haarnoja2018soft,
  title={Soft actor-critic: Off-policy maximum entropy deep reinforcement learning with a stochastic actor},
  author={Haarnoja, Tuomas and Zhou, Aurick and Abbeel, Pieter and Levine, Sergey},
  booktitle={International conference on machine learning},
  pages={1861--1870},
  year={2018},
  organization={Pmlr}
}

@article{schulman2017proximal,
  title={Proximal policy optimization algorithms},
  author={Schulman, John and Wolski, Filip and Dhariwal, Prafulla and Radford, Alec and Klimov, Oleg},
  journal={arXiv preprint arXiv:1707.06347},
  year={2017}
}

@inproceedings{tavakoli2018action,
  title={Action branching architectures for deep reinforcement learning},
  author={Tavakoli, Arash and Pardo, Fabio and Kormushev, Petar},
  booktitle={Proceedings of the aaai conference on artificial intelligence},
  volume={32},
  number={1},
  year={2018}
}

@inproceedings{wu2025rlvr,
  title={RLVR-World: Training World Models with Reinforcement Learning},
  author={Wu, Jialong and Yin, Shaofeng and Feng, Ningya and Long, Mingsheng},
  booktitle={The Thirty-ninth Annual Conference on Neural Information Processing Systems}
}

@inproceedings{georgiev2025pwm,
  title={PWM: Policy Learning with Multi-Task World Models},
  author={Georgiev, Ignat and Giridhar, Varun and Hansen, Nicklas and Garg, Animesh},
  booktitle={The Thirteenth International Conference on Learning Representations}
}

@inproceedings{zhao2025generalist,
  title={Generalist World Model Pre-Training for Efficient Reinforcement Learning},
  author={Zhao, Yi and Scannell, Aidan and Hou, Yuxin and Cui, Tianyu and Chen, Le and B{\"u}chler, Dieter and Solin, Arno and Kannala, Juho and Pajarinen, Joni},
  booktitle={ICLR 2025 Workshop on World Models: Understanding, Modelling and Scaling},
  year={2025}
}

@article{hafner2025mastering,
  title={Mastering diverse control tasks through world models},
  author={Hafner, Danijar and Pasukonis, Jurgis and Ba, Jimmy and Lillicrap, Timothy},
  journal={Nature},
  volume={640},
  number={8059},
  pages={647--653},
  year={2025},
  publisher={Nature Publishing Group UK London}
}

@article{bardes2024vjpea,
  title={Revisiting Feature Prediction for Learning Visual Representations from Video},
  author={Bardes, Adrien and Garrido, Quentin and Ponce, Jean and Chen, Xinlei and Rabbat, Michael and LeCun, Yann and Assran, Mido and Ballas, Nicolas},
  journal={Transactions on Machine Learning Research}
}

@inproceedings{bruce2024genie,
  title={Genie: Generative interactive environments},
  author={Bruce, Jake and Dennis, Michael D and Edwards, Ashley and Parker-Holder, Jack and Shi, Yuge and Hughes, Edward and Lai, Matthew and Mavalankar, Aditi and Steigerwald, Richie and Apps, Chris and others},
  booktitle={Forty-first International Conference on Machine Learning},
  year={2024}
}

@inproceedings{lu2024rethinking,
  title={Rethinking Transformers in Solving POMDPs},
  author={Lu, Chenhao and Shi, Ruizhe and Liu, Yuyao and Hu, Kaizhe and Du, Simon Shaolei and Xu, Huazhe},
  booktitle={International Conference on Machine Learning},
  pages={33089--33112},
  year={2024},
  organization={PMLR}
}

@inproceedings{zengpoliformer,
  title={PoliFormer: Scaling On-Policy RL with Transformers Results in Masterful Navigators},
  author={Zeng, Kuo-Hao and Zhang, Zichen and Ehsani, Kiana and Hendrix, Rose and Salvador, Jordi and Herrasti, Alvaro and Girshick, Ross and Kembhavi, Aniruddha and Weihs, Luca},
  booktitle={8th Annual Conference on Robot Learning}
}

@article{ni2023transformers,
  title={When do transformers shine in rl? decoupling memory from credit assignment},
  author={Ni, Tianwei and Ma, Michel and Eysenbach, Benjamin and Bacon, Pierre-Luc},
  journal={Advances in Neural Information Processing Systems},
  volume={36},
  pages={50429--50452},
  year={2023}
}

@article{xu2024integrating,
  title={Integrating mamba and transformer for long-short range time series forecasting},
  author={Xu, Xiongxiao and Liang, Yueqing and Huang, Baixiang and Lan, Zhiling and Shu, Kai},
  journal={arXiv preprint arXiv:2404.14757},
  volume={2},
  number={7},
  year={2024}
}

@article{sun2023building,
  title={Building occupancy number prediction: A Transformer approach},
  author={Sun, Kailai and Qaisar, Irfan and Khan, Muhammad Arslan and Xing, Tian and Zhao, Qianchuan},
  journal={Building and environment},
  volume={244},
  pages={110807},
  year={2023},
  publisher={Elsevier}
}

@article{kim2019real,
  title={Real-time occupancy prediction in a large exhibition hall using deep learning approach},
  author={Kim, Seonghyeon and Kang, Seokwoo and Ryu, Kwang Ryel and Song, Giltae},
  journal={Energy and Buildings},
  volume={199},
  pages={216--222},
  year={2019},
  publisher={Elsevier}
}

@article{dong2026optimal,
  title={Optimal energy management of buildings using neural network-based thermal prediction and economic model predictive control},
  author={Dong, Zihang and Hu, Cheng and Zhang, Xi and Shen, Yifan and Shen, Xiaojun and Leon, Jose I},
  journal={Advanced Engineering Informatics},
  volume={71},
  pages={104278},
  year={2026},
  publisher={Elsevier}
}

@article{ren2026agenticontrol,
  title={AgentiControl-MPC: Automated model predictive control for buildings using large language model-based multi-agent system},
  author={Ren, Chenxiao and Hong, Sung-Min and Tang, Rui},
  journal={Advanced Engineering Informatics},
  volume={76},
  pages={105062},
  year={2026},
  publisher={Elsevier}
}

@article{luo2019development,
  title={Development of an IoT-based big data platform for day-ahead prediction of building heating and cooling demands},
  author={Luo, XJ and Oyedele, Lukumon O and Ajayi, Anuoluwapo O and Monyei, Chukwuka G and Akinade, Olugbenga O and Akanbi, Lukman A},
  journal={Advanced Engineering Informatics},
  volume={41},
  pages={100926},
  year={2019},
  publisher={Elsevier}
}

@article{zhang2026physics,
  title={Physics-informed band structure-integrated Continuous-time inhomogeneous Markov chains for stochastic occupancy modeling},
  author={Zhang, Hanbei and Thilker, Christian Ankerstjerne and Xiao, Fu and Madsen, Henrik and Li, Rongling and Ma, Tianyou and Xu, Kan},
  journal={Advanced Engineering Informatics},
  volume={71},
  pages={104204},
  year={2026},
  publisher={Elsevier}
}

@article{glashier2024iterative,
  title={An iterative regression-based thermal response prediction methodology for instrumented civil infrastructure},
  author={Glashier, Theo and Kromanis, Rolands and Buchanan, Craig},
  journal={Advanced Engineering Informatics},
  volume={60},
  pages={102347},
  year={2024},
  publisher={Elsevier}
}

@article{zhao2014dynamic,
  author  = {Zhao, Kang and Liu, Xiao-Hua and Jiang, Yi},
  title   = {Dynamic performance of water-based radiant floors during
             start-up and high-intensity solar radiation},
  journal = {Solar Energy},
  year    = {2014},
  volume  = {101},
  pages   = {232--244},
  doi     = {10.1016/j.solener.2013.12.033}
}


\bio{}

\endbio

\end{document}